\documentclass[10pt,journal,compsoc]{IEEEtran}
\usepackage{amsmath,amsfonts}
\usepackage{algorithmic}
\usepackage{algorithm}
\usepackage{array}
\usepackage[caption=false,font=normalsize,labelfont=sf,textfont=sf]{subfig}
\usepackage{textcomp}
\usepackage{stfloats}
\usepackage{url}
\usepackage{verbatim}
\usepackage{graphicx}
\usepackage{multirow}
\usepackage[table,xcdraw]{xcolor}
\usepackage{booktabs}

\usepackage{algorithm}
\usepackage{algorithmic}
\usepackage{amssymb}
\usepackage{amsmath}
\usepackage{diagbox}

\definecolor{boxhead}{rgb}{0.70196078, 0.80392157, 0.89019608}
\definecolor{boxbody}{rgb}{0.9215, 0.9215, 0.9215}
\definecolor{boxus}{rgb}{0.9961, 0.8510, 0.6510}

\usepackage{xcolor}

\newcommand{\todo}[1]{{\color{red}{\bf [#1]}}}

\newcommand{\etal}{{\textit{et al.}}}
\newcommand{\ie}{{\textit{i.e.}}}
\newcommand{\eg}{{\textit{e.g.}}}
\newcommand{\nobf}[1]{\noindent{\bf #1}}

\usepackage{newfloat}
\usepackage{listings}

\usepackage{caption}

\usepackage{gensymb}
\usepackage{pifont}

\ifCLASSOPTIONcompsoc
  \usepackage[nocompress]{cite}
\else
  \usepackage{cite}
\fi
\ifCLASSINFOpdf
\else
\fi
\definecolor{cvprblue}{rgb}{0,0,1}
\usepackage[pagebackref,breaklinks,colorlinks,citecolor=cvprblue]{hyperref}

\title{OCMG-Net: Neural Oriented Normal Refinement for Unstructured Point Clouds}

\usepackage{bibentry}

\begin{document}

\title{OCMG-Net: Neural Oriented Normal Refinement for Unstructured Point Clouds} 
\author{Yingrui Wu,
Mingyang~Zhao,
Weize Quan, 
Jian Shi,
Xiaohong~Jia,
Dong-Ming Yan
\IEEEcompsocitemizethanks{
\IEEEcompsocthanksitem Y. Wu, J. Shi, W. Quan, and D. Yan are with the Institute of Automation, CAS, Beijing, China. E-mail: \{wuyingrui2023, jian.shi\}@ia.ac.cn, \{qweizework, yandongming\}@gmail.com.\protect
\IEEEcompsocthanksitem M. Zhao is with the Hong Kong Institute of Science \& Innovation Chinese Academy of Sciences, CAS, HongKong, China. E-mail: migyangz@gmail.com.\protect
\IEEEcompsocthanksitem X. Jia, is with the Academy of Mathematics and Systems Science, CAS, Beijing, China. E-mail: xhjia@amss.ac.cn.\protect
}
}
	
\IEEEtitleabstractindextext{
\begin{center}\setcounter{figure}{0}
\centering
\includegraphics[width=0.9\linewidth]{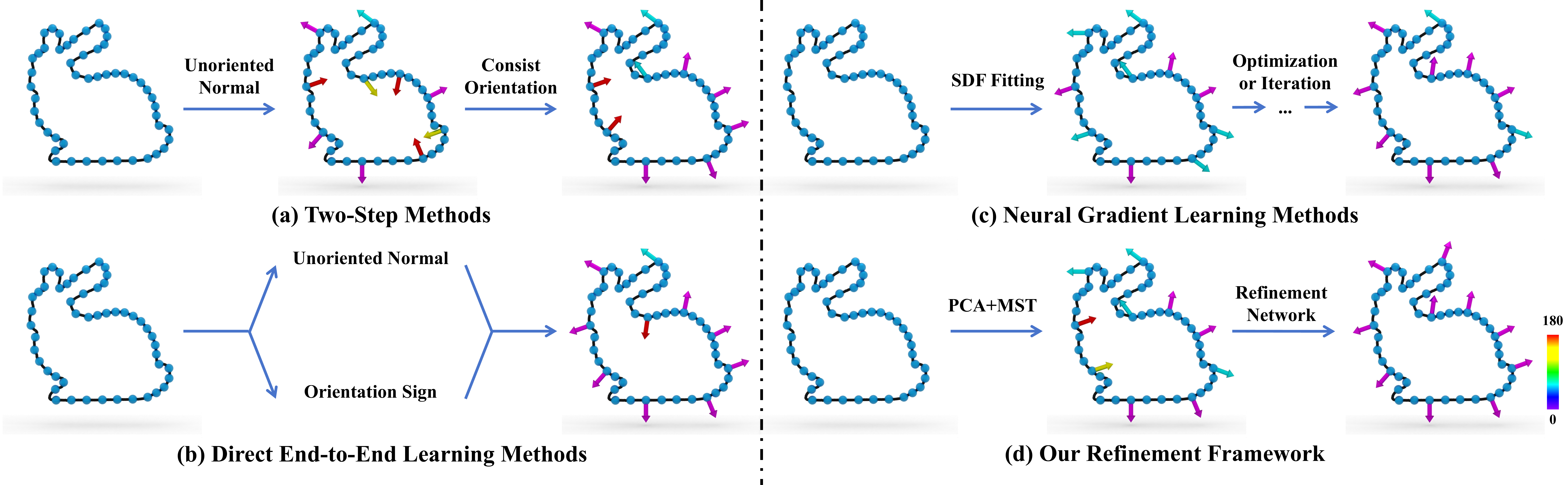}
\captionof{figure}{The 2D schematic representation of the four principal oriented normal estimation manners.}
\label{fig: orientation manner}
\end{center}

\begin{abstract}

We present a robust refinement method for estimating oriented normals from unstructured point clouds. In contrast to previous approaches that either suffer from high computational complexity or fail to achieve desirable accuracy, our novel framework incorporates sign orientation and data augmentation in the feature space to refine the initial oriented normals, striking a balance between efficiency and accuracy. To address the issue of noise-caused direction inconsistency existing in previous approaches, we introduce a new metric called the \emph{Chamfer Normal Distance}, which faithfully minimizes the estimation error by correcting the annotated normal with the closest point found on the potentially clean point cloud. This metric not only tackles the challenge but also aids in network training and significantly enhances network robustness against noise. Moreover, we propose an innovative dual-parallel architecture that integrates \emph{Multi-scale Local Feature Aggregation} and \emph{Hierarchical Geometric Information Fusion}, which enables the network to capture intricate geometric details more effectively and notably reduces ambiguity in scale selection. Extensive experiments demonstrate the superiority and versatility of our method in both unoriented and oriented normal estimation tasks across synthetic and real-world datasets among indoor and outdoor scenarios. The code is available at \url{https://github.com/YingruiWoo/OCMG-Net.git}.

\end{abstract}
}

\maketitle

\IEEEdisplaynontitleabstractindextext

\section{Introduction}\label{sec:intro}
\subsection{Background}
\IEEEPARstart{N}{ormal} estimation is a fundamentally important task in the field of \emph{point cloud analysis}. Normals with consistent orientation have a wide variety of applications in 3D vision and robotics, such as graphic rendering~\cite{bui1975illumination,santo2020deep,wimbauer2022rendering}, surface reconstruction~\cite{fleishman2005robust,kazhdan2006poisson,ben2022digs}, and semantic segmentation~\cite{grilli2017review,che2018multi}. In recent years, many powerful methods have been developed to enhance the performance of normal estimation. However, most of them merely return unoriented normals, where further orientation step is needed. In the context of \emph{oriented} normal estimation, most approaches often fail to \emph{balance accuracy and efficiency}. Moreover, the latest methods involving both traditional and learning-based ones still suffer from \emph{noise disturbances} and struggle to attain high-quality results for point clouds with \emph{complex geometries}.

Conventionally, the estimation of oriented normals is performed in two steps, as shown in Fig.~\ref{fig: orientation manner}(a): \emph{unoriented normal estimation} and \emph{normal orientation}. In the first step, traditional~\cite{hoppe1992surface,levin1998approximation,cazals2005estimating} and deep surface fitting methods~\cite{lenssen2020deep,ben2020deepfit,zhu2021adafit,li2022graphfit,du2023rethinking} typically encompass fitting local planes or polynomial surfaces and then infer normal vectors from the fitted outcomes. Nevertheless, these approaches face challenges when generalizing to complex shapes. Moreover, their performance heavily relies on parameter tuning. Benefiting from the enhanced feature extraction capabilities, recent regression-based methods~\cite{guerrero2018pcpnet,ben2019nesti,zhou2020geometry,wang2020neighbourhood,cao2021latent,zhou2022fast,zhang2022geometry,li2022hsurf,xiu2023msecnet} have advanced normal estimation for clean point clouds, however, they have not made significant progress in handling normal estimation for noisy point clouds. When it comes to normal orientation, most methods~\cite{hoppe1992surface, konig2009consistent, schertler2017towards, xu2018towards, jakob2019parallel, metzer2021orienting} adopt a greedy propagation strategy based on the \emph{local consistency hypothesis}, \ie, points within a local patch have similar normal vectors. As a result, the tuning of the propagation neighbor size is crucial, and errors in local areas can propagate to subsequent steps. Due to the assumption of local consistency, even if accurate unoriented normals are obtained, such as those preserving sharp edges, the resulting orientations may still not be optimal.

Recently, learning-based approaches for oriented normal estimation try to integrate the previously separate two steps into an end-to-end manner. As presented in Fig.~\ref{fig: orientation manner}(b), the most straightforward methods directly estimate the oriented normal~\cite{guerrero2018pcpnet,hashimoto2019normal} or predict the unoriented normal and orient the sign (\ie, $\pm$)~\cite{wang2022deep,li2023shs,li2024learning} based on a shared feature. However, these methods heavily rely on the feature representation of geometric structures, making it challenging to achieve efficient estimation with high accuracy. In contrast, approaches based on neural gradient learning~\cite{li2023neural, li2024neuralgf} as depicted in Fig.~\ref{fig: orientation manner}(c), employ implicit neural representations to fit the underlying surface of individual point clouds. They use the gradient vectors of the \emph{Signed Distance Function} (SDF)~\cite{mescheder2019occupancy,park2019deepsdf} to represent the oriented normals. 
While more accurate orientation results are achieved, the implicit neural representation needs to be re-trained from scratch for each new point cloud. Moreover, additional optimization steps or iterative strategies are necessary to refine the final outcomes, leading to extended computational cost.


\subsection{Motivation and the Proposed Solution}
To address the aforementioned issues in oriented normal estimation, including \emph{balancing accuracy and efficiency}, \emph{enhancing robustness to noise}, and \emph{obtaining effective feature descriptions for intricate geometries}, we present a novel framework paired with a multi-scale geometry enhanced network architecture for oriented normal estimation.
Our method offers the following advantages: 1) An end-to-end refinement framework that provides more accurate estimations and strikes a better balance between accuracy and efficiency.
2) Introduction of the novel \emph{Chamfer Normal Distance} (CND) metric to resolve discrepancies between annotated normals and the neighbor geometries of noisy points.
3) A network that integrates diverse geometric information extraction within a hierarchical framework to enhance the effective capture of complex geometries.

Given the limited sampling sizes of unstructured point clouds, it is significantly challenging for end-to-end approaches to accurately determine orientation details such as holes and gaps based on the learned descriptions of the sparse inputs. Additionally, end-to-end methods directly employ a binary classification approach for estimating orientation signs, which oversimplifies the orientation problem and may struggle to the diverse orientation scenarios. Inspired by the prevalent use of \emph{Principal Component Analysis} (PCA) in initializing input patches for unoriented normal learning methods, we extend this approach to oriented normal estimation to effectively simplify the input data. Specifically, as presented in Fig.~\ref{fig: orientation manner}(d), we combine PCA with \emph{Minimum Spanning Tree} (MST) to approximate an initial oriented normal. This step essentially transforms the direct end-to-end process into a \emph{soft} refinement framework. With the guidance of the initialized normal, it becomes easier for networks to identify failures in MST rather than directly solving the \emph{hard} binary classification problem. Additionally, for a more informed feature description, we conduct orientation sign refinement in the feature space alongside effective data augmentation. Leveraging the low computational cost of MST and the proposed data augmentation scheme, our framework substantially enhances both the accuracy and efficiency of the end-to-end approach.


\begin{figure}[t]
\centering
\includegraphics[width=\linewidth]{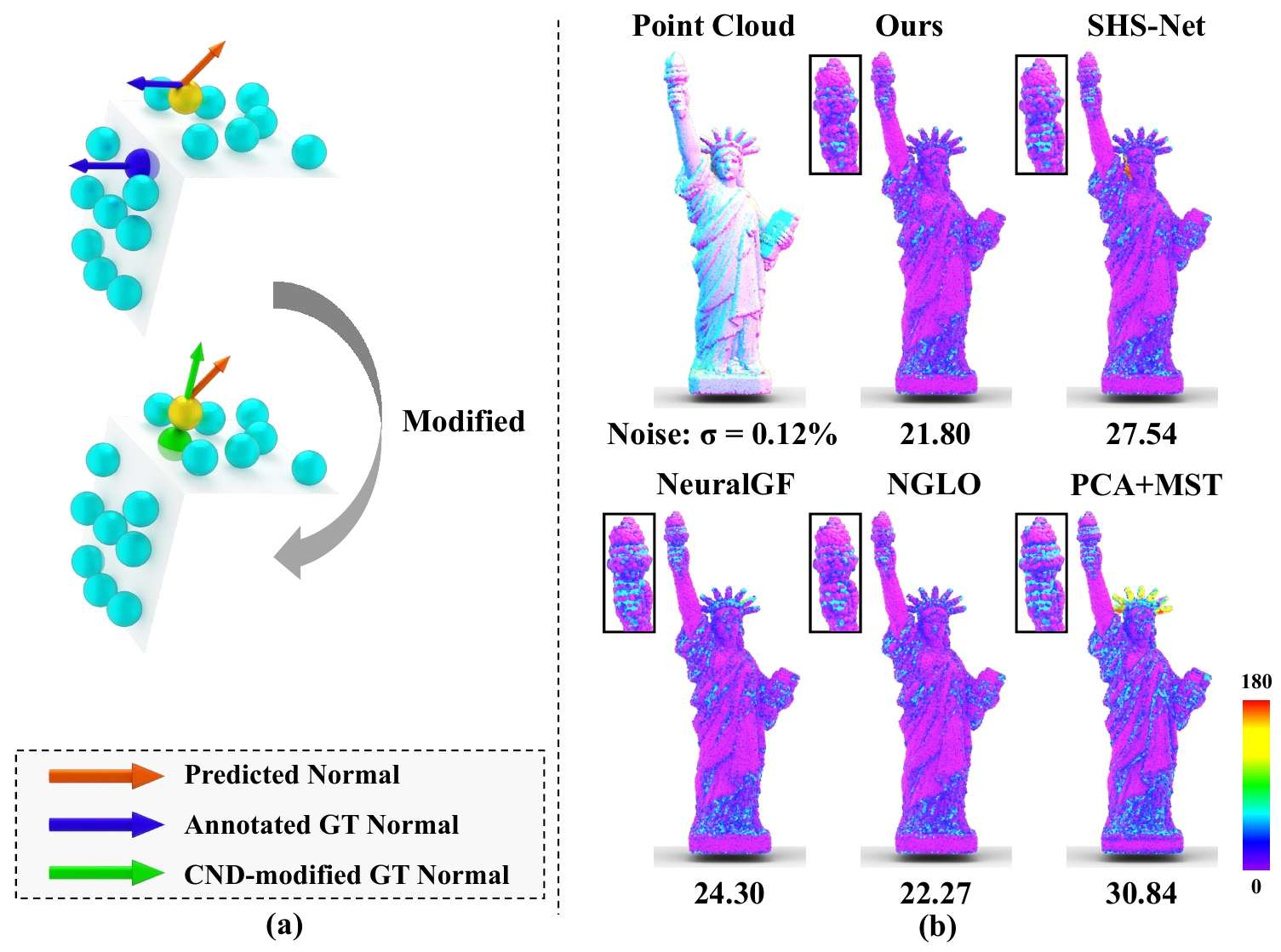}
\vskip -0.2cm
\caption{(a) A comparison between the annotated and the proposed CND-modified ground-truth (GT) normals, where the latter is more consistent with the underlying surface geometries. (b) Our method outperforms competitors with higher robustness to noise and intricate shape details, as highlighted by the heat map.}
\label{fig: methods_comparison}
\vskip -0.5cm
\end{figure}
In our oriented normal estimation framework, we delve into a further analysis of the normal estimation deviations caused by noisy point clouds. 
We identify the \emph{inconsistency} between the annotated normal and the input patch, as illustrated in Fig.~\ref{fig: methods_comparison}(a). We find that this direction inconsistency significantly affects both network training and output evaluation. The reason is that when point coordinates undergo significant changes due to noise disturbances, their neighbor geometries and normals consequently change as well, while the annotated normals remain constant. 
To tackle this challenge, we introduce a more appropriate metric, CND, for normal estimation in place of the conventional \emph{Root of Mean Squared Error} (RMSE). CND substitutes the original annotated normal vector with the normal of the nearest point found on the potentially clean point cloud. By incorporating CND into the loss function, we effectively mitigate disturbances stemming from inconsistency deviations during training. Moreover, we experimentally validate that our newly defined loss function attains superior normal estimation accuracy compared to competing approaches.



Additionally, we combine the CND-modified loss function with multi-scale geometric structures to achieve more stable and robust oriented normal estimation. Previous approaches often struggle to extract multi-scale geometric information effectively, facing challenges related to scale ambiguity between noise smoothing and detail preservation. To address this, we employ a multi-scale local feature extraction strategy, enhanced by integration through an \emph{attention} layer, to learn a much more resilient cross-scale feature. This significantly enriches individual point features with intricate geometric details and reduces noise disturbances. Subsequently, we employ a hierarchical architecture with cross-scale global features to enhance multi-scale global information for network inference. In the context of oriented normal estimation, our proposed architecture involves \emph{a dual-parallel feature extraction structure} to capture comprehensive descriptions of the input patch and subsampled point cloud.

We conduct extensive experiments to validate our proposed method and compare it with the latest approaches across various benchmark datasets. The results demonstrate that our method outperforms the baselines by a significant margin in \emph{both unoriented} and \emph{oriented} normal estimation tasks. Especially, our method excels in scenarios involving noisy point clouds, intricate geometric details, and varying distribution density. Moreover, as demonstrated empirically, the oriented normals estimated by our method bring superior \emph{reconstruction} results  than competitors. 

This journal paper is an extended and enhanced version of our preliminary AAAI conference paper~\cite{wu2024cmg}. We extend the previous work from 
unoriented normal estimation to the \emph{oriented field} and delve much deeper into our analysis from both \emph{theoretical} and \emph{experimental} aspects. Moreover, we make substantial additional improvements and refinements:

\begin{itemize}
    \item We introduce a novel framework that generalizes the prior unoriented normal estimation network to an oriented one termed OCMG-Net. Our framework can be seamlessly adapted in other end-to-end approaches to enhance their efficacy. This flexibility and versatility makes our method a valuable tool for a range of existing methods within the field.
    \item We conduct an in-depth theoretical analysis of the proposed CND metric from both unoriented and oriented normal estimation viewpoints. Meanwhile, we experimentally validate the efficacy of the CND-modified loss. The application of the CND metric allows us to establish a more rational benchmark for evaluating the performance of existing normal estimation approaches.


    \item   We develop a dual-parallel structure to learn comprehensive descriptions of the inputs patch and point cloud and empirically demonstrate that with the proposed architecture, OCMG-Net can simultaneously handle complex geometries, density variations, and different levels of noise within the data. Moreover, OCMG-Net exhibits strong generalization abilities when confronted with data from various sources.

    \item 
    We perform extensive 
    experiments on more intricate and diverse datasets, encompassing synthetic and real-world data from both indoor and outdoor scenes with the latest methods. Experimental results demonstrate that our method advances the state-of-the-art performance in both unoriented and oriented normal estimation tasks.

\end{itemize}

\section{Related Work}
\label{rw}
In this section, we provide an overview of previous approaches to normal estimation from both unoriented and orientation perspectives. 

\subsection{Unoriented Normal Estimation}
Over the past years, point cloud normal estimation has been an active research due to its wide-ranging applications across various domains. The estimation process is typically divided into two main steps: \emph{unoriented normal estimation} and \emph{normal orientation}.

In the initial unoriented normal estimation step, PCA~\cite{hoppe1992surface} has emerged as one of the most commonly used methods, involving the fitting of a plane to the input point cloud patch. Subsequent methods have introduced variations on PCA\cite{alexa2001point,pauly2002efficient,mitra2003estimating,lange2005anisotropic,huang2009consolidation} and more complex surface fitting techniques~\cite{levin1998approximation,cazals2005estimating,guennebaud2007algebraic,aroudj2017visibility,oztireli2009feature}. These methods aim to reduce the noise influence by selecting larger patches and fitting the underlying surface more accurately through complex functions such as spheres and high-order polynomials. However, they often tend to oversmooth sharp features and intricate geometric details. To handle these challenges, techniques such as Voronoi diagrams~\cite{amenta1998surface,dey2004provable,alliez2007voronoi,merigot2010voronoi}, Hough transform~\cite{boulch2012fast}, and plane voting~\cite{zhang2018multi} have been deployed. Nevertheless, these methods frequently require manual parameter tuning, constraining their practical applicability and scalability in real-world scenarios.

With the rapid advancement of neural networks, learning-based approaches for normal estimation have shown improved performance and reduced dependence on manual parameter tuning compared to traditional ones. These learning-based approaches can be broadly categorized into two main groups: \emph{deep surface fitting} and \emph{regression-based methods}. Deep surface fitting approaches~\cite{lenssen2020deep, ben2020deepfit,zhu2021adafit,li2022graphfit,du2023rethinking} utilize deep neural networks to predict point-wise weights, subsequently fitting a polynomial surface to input point cloud patches through \emph{weighted least-squares} optimization. However, deep surface fitting methods may face challenges like overfitting or underfitting due to the fixed order of the objective polynomial functions.

On the other hand, regression-based methods~\cite{boulch2016deep,lu2020deep,guerrero2018pcpnet,qi2017pointnet,hashimoto2019normal,ben2019nesti,zhou2022refine,zhang2022geometry,li2022hsurf,li2023neaf,li2023shs,li2024learning,xiu2023msecnet} formulate the normal estimation task as a regression problem based on features extracted by point cloud processing networks like PointNet~\cite{qi2017pointnet} and PointNet++~\cite{qi2017pointnet++}. Leveraging the robust feature extraction capabilities of these network architectures, recent regression-based methods~\cite{li2022hsurf,li2023neaf,li2023shs,li2024learning,xiu2023msecnet} have shown promising results on clean point clouds. For instance, Hsurf-Net~\cite{li2022hsurf} and SHS-Net~\cite{li2023shs,li2024learning} optimized a hyper surface description for the inputs and parameterized the normal fitting by \emph{Multi-Layer Perception} (MLP). To learn more geometric information, NeAF~\cite{li2023neaf} proposed to predict the angle offsets of query vectors. MSECNet~\cite{xiu2023msecnet} improved normal estimation in sharp areas through the incorporation of a novel edge detection technology. Nevertheless, challenges remain in normal estimation for noisy point clouds, which constrains their practical utility.

\subsection{Normal Orientation}
The unoriented normals estimated by the aforementioned methods require further consistent orientation for downstream applications. Utilizing a greedy propagation strategy, the widely used work~\cite{hoppe1992surface} and its variants~\cite{konig2009consistent,seversky2011harmonic,wang2012variational,schertler2017towards,xu2018towards,jakob2019parallel} propagated the normal orientation of seed points to their adjacent points via an MST. Recent work by Metze~\etal~\cite{metzer2021orienting} introduce a dipole propagation strategy across partitioned patches to achieve global consistency. Nevertheless, these methods heavily rely on tuning the propagation neighbor size. Moreover, due to the assumption of local consistency, errors in local areas may persist in subsequent steps, and sharp features are often not well-preserved. In contrast to propagation methods, alternatives involve various shape representation methodologies, such as  SDF \cite{mello2003estimating,mullen2010signing}, variational formulations \cite{walder2005implicit,alliez2007voronoi,huang2019variational}, visibility considerations~\cite{katz2007direct,chen2010binary}, isovalue constraints~\cite{xie2004surface}, active contour methods~\cite{xiao2023point}, and winding-number field~\cite{liu2024consistent,xu2023globally}. Although these approaches partially overcome the limitations of local consistency, the continuity of representations still poses challenges in preserving sharp edges. 

Recently, several learning-based approaches have sought to streamline the split oriented normal estimation process into a single step to achieve higher accuracy. The most straightforward end-to-end approach estimate the oriented normals~\cite{guerrero2018pcpnet,hashimoto2019normal} directly. In contrast, based on both global point cloud and local patch features, other techniques predict the unoriented normal and its orientation sign separately~\cite{wang2022deep,li2023shs,li2024learning}, essentially transforming consistent orientation into a hard \emph{binary classification} problem for each unoriented normal.  More recently, neural gradient learning methods~\cite{li2023neural,li2024neuralgf} utilize neural networks to model the SDF of individual point clouds and leverage gradient vectors to define the oriented normals. While these methods enhance accuracy, they necessitate training from scratch for each point cloud and commonly incorporate additional optimization steps~\cite{li2023neural} or iterative strategies~\cite{li2024neuralgf}, thereby increasing computational costs.

\section{Rethinking Oriented Normal Estimation}
\label{re}
To balance both accuracy and  efficiency  in oriented normal estimation, we first analyze the reasons behind the limitations of both end-to-end and neural gradient learning methods and find that a proper initialization can significantly enhance the generalizability of the end-to-end approach with minimal additional computational time, thereby establishing an efficient refinement framework. Moreover, to enhance performance on noisy point clouds with complex geometries, we identify the direction inconsistency induced by noise and deliberate on solutions for resolving scale ambiguity in feature extraction.


\begin{figure}[t]
\centering
\includegraphics[width=\linewidth]{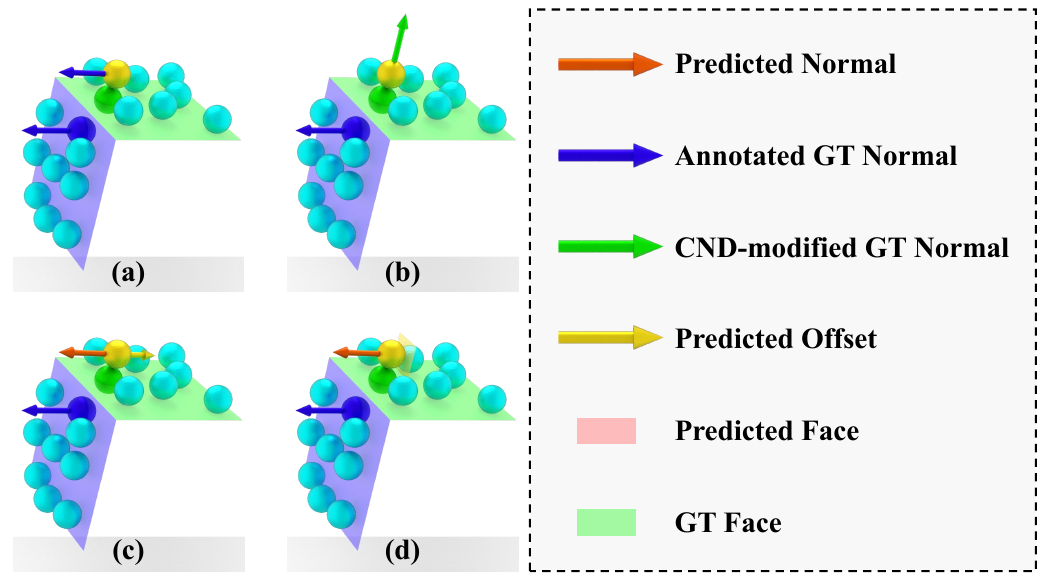}
\caption{(a) The annotated ground-truth normal $\boldsymbol{n}_{\boldsymbol{p}_i}$ of the noisy point $\boldsymbol{p}_i$ determined before noisy disturbance indeed is inconsistent with the input patch. (b) The direction of the normal $\boldsymbol{n}_{\tilde{\boldsymbol{p}}_i}$ of the nearest clean point $\tilde{\boldsymbol{p}}_i$ is more consistent with the input patch, which is taken as the CND-modified GT normal. (c) The predicted offset $\hat{\boldsymbol{d}}_{\boldsymbol{p}_i}$ cannot drag $\boldsymbol{p}_i$ to the noise-free underlying surface. (d) This inconsistency also degrades the surface reconstruction performance. }
\label{fig: comparison_between_the_ground-truth_normals}
\vskip -0.5cm
\end{figure}

\subsection{Balance between Accuracy and Efficiency}
The recent end-to-end approaches for oriented normal estimation simplify the orientation into a binary classification task, predicting the unoriented normal and the orientation sign based on a shared feature, separately. However, these methods often suffer from overfitting and struggle with generalizing the orientation sign prediction due to the inherent randomness from unoriented normal estimation and oversimplified orientation problem. Specifically, the sparse nature of the sampled point cloud, stemming from limited input size, fails to provide sufficient information for detailed sign orientation. Moreover, the original orientations of estimated unoriented normals rely solely on the input local patches, leading to a distribution gap in the binary classification of orientation sign when encountering various global point clouds. Additionally, predicting two distinct geometric attributes utilizing a shared feature tends to constrain their individual performance. In contrast, neural gradient learning methods fit the SDF to each point cloud and derive the oriented normal from the gradient of the SDF, obviating the need for sign classification. However, training the neural SDF for each point cloud necessitates a distinct training process. The SDF fitting, coupled with additional optimization and iteration steps, greatly extends computational complexity.


We reevaluate previous approaches and facilitate the end-to-end manner into an efficient and accurate method. We notice that PCA is commonly utilized in unoriented normal estimation techniques to rotate the input patch, essentially serving as an unoriented normal initialization. We also discover that neural networks find it easier to enhance the initialized normal estimation outcomes rather than directly estimating the normal from the original input patch. As for oriented normal estimation,  adhering to the smoothness assumption, the normal predicted by PCA can be assigned an initial sign by MST. Leveraging the prior information provided by MST, we find that refining the MST error achieves better results than directly classifying the sign of a random unoriented normal. Within this framework, we further decouple the features of unoriented normal estimation and consistent orientation, executing sign refinement and data augmentation in the feature space to enhance both efficacy and generalization capabilities. Thanks to its minimal time overhead, the proposed framework efficiently improves the accuracy of the end-to-end framework.


\subsection{Direction Inconsistency}
Subsequently, we further analyze the shortcomings of the existing approaches in noisy point cloud normal estimation. Previous learning-based approaches directly minimize the deviations between the predicted normals and the annotated ones for training and evaluation. This is reasonable for noise-free scenarios, however, for noisy point clouds, due to the noise-caused relative coordinate changes, the annotated normals indeed are inconsistent with the neighbor geometries of the query points. As presented in Fig.~\ref{fig: comparison_between_the_ground-truth_normals}(a), given a set of noisy point clouds $\mathcal{P}$, suppose the ground-truth projection locating on the noise-free surface of the noisy point $\boldsymbol{p}_i$ is $\tilde{\boldsymbol{p}}_i$. As depicted in Figs.~\ref{fig: comparison_between_the_ground-truth_normals}(a) and (b), the annotated normal of $\boldsymbol{p}_i$ is $\boldsymbol{{n}}_{\boldsymbol{p}_i}\in \mathbb{R}^3$, which is the same as the one of the point before adding noise, and the normal of $\tilde{\boldsymbol{p}}_i$ is $\boldsymbol{n}_{\tilde{\boldsymbol{p}}_i}\in \mathbb{R}^3$. If we optimize the conventionally defined normal estimation loss $\|\boldsymbol{n}_{\boldsymbol{p}_i}-\hat{\boldsymbol{{n}}}_{\boldsymbol{p}_i}\|_2^2$ as done by predecessors, where  $\hat{\boldsymbol{n}}_{\boldsymbol{p}_i}$ is the predicted normal, this will inevitably lead to inconsistency between the annotated normal $\boldsymbol{{n}}_{\boldsymbol{p}_i}$ and the input patch $\boldsymbol{P}_i$. What's worse, with the potential normal deviations and opposite orientation signs, this inconsistency greatly decreases the quality of the training data and thus lowers down the network's ability to estimate oriented normals accurately, particularly on noisy point clouds.


Moreover, this inconsistency also degrades downstream tasks such as denoising and 3D reconstruction. For instance, Fig.~\ref{fig: comparison_between_the_ground-truth_normals}(c) shows
the denosing principle for point clouds. If we utilize the predicted normal vector $\hat{\boldsymbol{n}}_{\boldsymbol{p}_i}$, which closely resembles the annotated normal vector $\boldsymbol{n}_{\boldsymbol{p}_i}$ (indicating a highly accurate estimation), the introduced offset $\hat{\boldsymbol{d}}_{\boldsymbol{p}_i}$ will not align or bring $\boldsymbol{p}_i$ closer to the noise-free underlying surface. Anonymously, in the context of reconstruction tasks, as shown in Fig.~\ref{fig: comparison_between_the_ground-truth_normals}(d), the regenerated mesh face $\widehat{F}_i$ in relation to the normal vector $\hat{\boldsymbol{{n}}}_{\boldsymbol{p}_i}$ significantly deviates from the authentic mesh fact ${F}_i$.

\begin{figure*}[htbp]
\centering
\includegraphics[width=0.9\linewidth]{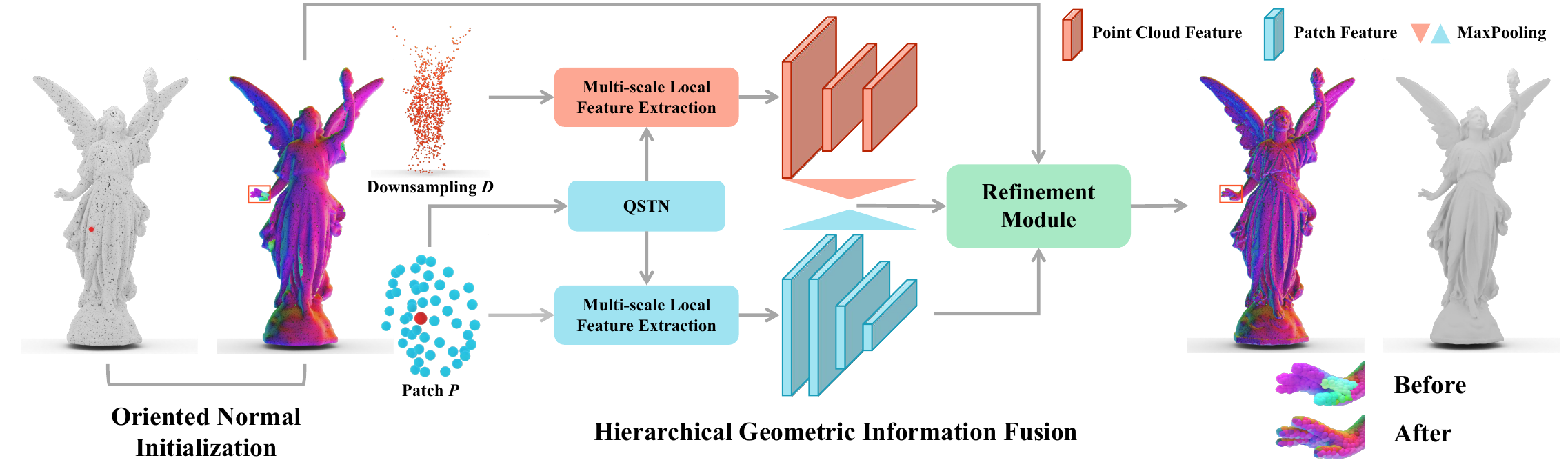}
\vskip -0.3cm
\caption{The schematic pipeline of the proposed OCMG-Net for orientated normal estimation. The initialized oriented normals acquired by PCA and MST are refined in the Refinement Module. This refinement process leverages features extracted from the dual-parallel Multi-scale Local Feature Extraction and Hierarchical Geometric Information Fusion approaches. Through this process, our method consistently delivers high-quality oriented normals, thereby ensuring accurate surface reconstruction.}
\label{fig: architecture_of_CMG-Net}
\vskip -0.3cm
\end{figure*}

\subsection{Scale Ambiguity}
In oriented normal estimation, the utilization of both patch and point cloud features is essential. However, a persistent challenge in existing approaches lies in the \emph{ambiguity} surrounding the optimal scale for both local and global feature extraction, particularly concerning patches.

Concerning local structures, employing large scales typically enhances noise robustness but can oversmooth shape details and sharp features. Conversely, small scales can preserve geometric nuances but are more susceptible to noise interference. In terms of global features, larger scales incorporate rich structural information from the underlying surface but may also introduce irrelevant points, potentially degrading the geometry details of the input patch. Conversely, smaller scales mitigate the inclusion of extraneous points but exhibit lesser resilience against noise.

Prior efforts suffer from challenges in efficiently extracting and integrating multi-scale local and global features, making them highly dependent on scale selection and resulting in unsatisfactory outcomes on both 
noisy point clouds and intricate shape details. Moreover, information loss during the forward path impedes the network from fully leveraging the distinct feature extraction scales. To enhance the network's feature description capabilities, we introduce weighted aggregation for multi-scale local features and integrate cross-scale information throughout the hierarchical process.

\section{The Proposed Method}

Based on the  aforementioned analysis, we propose a novel oriented normal estimation approach that is robust against noise and less sensitive to scale selection. Concrete technical contributions are presented in the following.

\subsection{Chamfer Normal Distance}

To bridge the direction inconsistency between the annotated normal and the geometry of the input patch, instead of using the conventional metric $\|\boldsymbol{n}_{\boldsymbol{p}_i}-\hat{\boldsymbol{n}}_{\boldsymbol{p}_i}\|_2^2$, inspired from the \emph{Chamfer Distance} (CD), 
we formulate the \emph{Chamfer Normal Distance} (CND) as
\begin{equation}
\text{CND}(\mathcal{P},{\tilde{\mathcal{P}}})=\sqrt{\frac{1}{N}\sum_{i=1}^{N}\mathrm{arccos}^2<\boldsymbol{n}_{\boldsymbol{\tilde{p}}_i}, \hat{\boldsymbol{n}}_{\boldsymbol{p}_i}>}, \nonumber
\end{equation}
where $<\cdot,\cdot>$ represents the inner product of two vectors and $\tilde{\boldsymbol{p}}_i$ is the closest point of $\boldsymbol{p}_i$ in the noise-free point cloud $\tilde{\mathcal{P}}$. In contrast to previous approaches that rely on annotated normal correspondence, our proposed CND manner assures consistency with the underlying geometric structure of the input patch (Fig.~\ref{fig: comparison_between_the_ground-truth_normals}(b)). The CND metric not only faithfully captures prediction errors within noisy point clouds, but also rectifies the direction inconsistency during network training, thereby significantly enhancing network robustness and facilitating subsequent application tasks.


\begin{figure}[t]
\centering
\includegraphics[width=\linewidth]{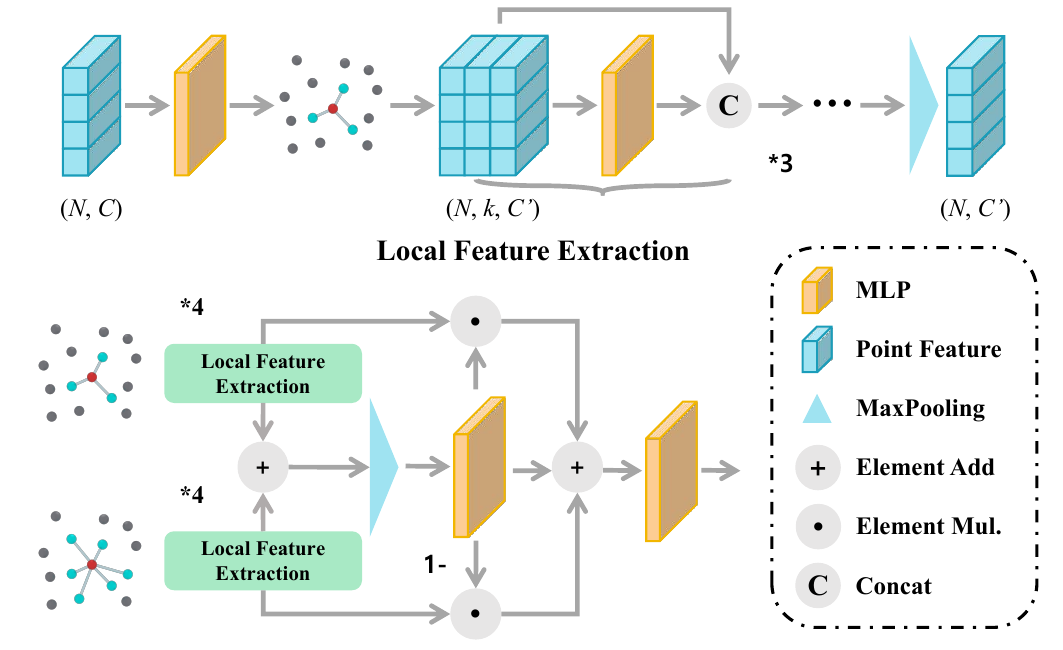}
\caption{The architecture of the proposed Multi-scale Local Feature Aggregation module.}
\label{fig: lfe}
\vskip -0.3cm
\end{figure}

\subsection{OCMG-Net}
To enhance the performance of the end-to-end manner, we utilize PCA and MST to initialize the oriented normal, thereby transitioning the end-to-end method into the subsequent refinement framework:
\begin{equation}
    \hat{\boldsymbol{n}}=\Psi_\mathrm{OCMG}(\boldsymbol{D}, \boldsymbol{P})\cdot\boldsymbol{n}_\mathrm{init}, \nonumber
\end{equation}where $\boldsymbol{n}_{init}$ represents the initialized oriented normal, $\Psi_{OCMG}$ is implemented by the neural network OCMG-Net, $\boldsymbol{P}=\{\boldsymbol{p}_i\in\mathbb{R}^3\}_{i=1}^{N_P}$ denotes a patch centered at a query point $\boldsymbol{p}$, and $\boldsymbol{D}=\{\boldsymbol{p}_i\in\mathbb{R}^3\}_{i=1}^{N_D}$ is a subset of the raw point cloud $\mathcal{P}$. The function $\Psi_{OCMG}$ can be used to predict both unoriented normals and orientation signs.  When considering a patch rotated by PCA, the unoriented normal estimated by OCMG-Net serves as a refinement over PCA, while in terms of orientation signs, the network can yield improved orientation outcomes based on the sign initialized by MST:
\begin{equation}
\hat{\boldsymbol{n}}=\Psi_{\boldsymbol{n}}(\boldsymbol{P}\mathbf{R})\mathbf{R}^{-1}\cdot\Psi_{s}(\boldsymbol{D}\mathbf{R}, \boldsymbol{P}\mathbf{R}, sgn_\mathrm{MST})\cdot{sgn}_\mathrm{MST}, \nonumber
\end{equation}
where $\Psi_{\boldsymbol{n}}$ and $\Psi_{s}$ represent the branches responsible for unoriented normal and orientation sign estimation within OCMG-Net, respectively. Here, $\mathbf{R}\in SO(3)$\footnote{$S O(3):=\left\{\mathbf{R} \in \mathbb{R}^{3 \times 3} \mid \mathbf{R} \mathbf{R}^{T}=\mathbf{R}^{T} \mathbf{R}=\mathbf{I}_{3}, \operatorname{det} \mathbf{R}=1\right\}.$} denotes the rotation matrix of PCA, and ${sgn}_\mathrm{MST}$ signifies the MST-initialized sign.

Moreover, we devise an architecture that integrates diverse geometric information extraction techniques within a \emph{hierarchical framework}. This design aims to capture a richer array of multi-scale structural information while addressing the challenge of scale ambiguity. Given the patch $\boldsymbol{P}$ centered at a query point $\boldsymbol{p}$ and the corresponding subset point cloud $\boldsymbol{D}$, as shown in Fig.~\ref{fig: architecture_of_CMG-Net}, OCMG-Net first normalizes the input points and rotates $\boldsymbol{P}$ and $\boldsymbol{D}$ by QSTN~\cite{qi2017pointnet, du2023rethinking} for better feature representation. Then, both $\boldsymbol{P}$ and $\boldsymbol{D}$ go through a parallel feature extraction network. In each branch, we group the local features by \emph{k-nearest neighbors} ($k$-NN) with different scales and aggregate them together, followed by a novel hierarchical structure with cross-scale geometry information fusion. Refined orientated normal estimation is conducted in the Refinement Module. For unoriented normal, position information fusion and weighted pooling are applied for to decode the embedding feature. When it comes to orientation sign, we project the MST-initialized sign to the normal feature space with a data augmentation and refine the MST result. Besides, our loss function modified by CND enables the network jumping out of the annotation inconsistency.

\begin{figure}[t]
\centering
\includegraphics[width=\linewidth]{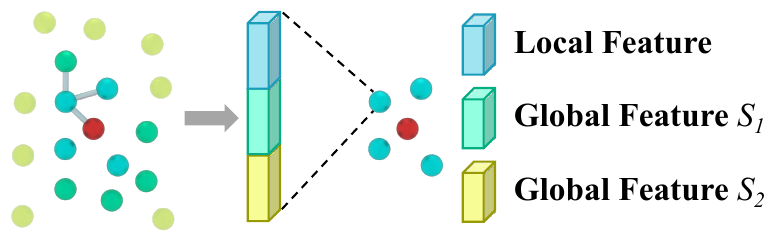}
\vskip -0.2cm
\caption{The fusion of cross-scale global features and local geometric information in the Hierarchical Geometric Information Fusion process.}
\label{fig: hgif}
\vskip -0.3cm
\end{figure}

\subsubsection{Multi-scale Local Feature Aggregation}
Graph-based local feature grouping is an effective way to enhance the description of each single point with its neighbor geometric structure~\cite{wang2019dynamic,li2022graphfit,li2022hsurf}. However, the widely used scale-fixed pooling manner often suffers from scale ambiguity and struggles to achieve an adaptive representation among various levels of noise and shape details. To solve this issue,  as presented in Fig.~\ref{fig: lfe}, we design a scale adaptation manner, consisting of cascade multi-scale \emph{Local Feature Extraction} blocks and an attention-based feature aggregation. The {Local Feature Extraction} block constructs graphs by $k$-NN with various scales and employs the skip-connection and maxpooling to capture the local structures, which can be formulated as
\begin{equation}
\fontsize{6.5pt}{\baselineskip}\selectfont
\boldsymbol{f}_{i}^{n+1}\!=\!\mathrm{MaxPool}\left\{\phi_1\left(\varphi_1\left(\boldsymbol{f}_{i}^{n}\right), \varphi_1\left(\boldsymbol{f}_{i, j}^{n}\right), \varphi_1\left(\boldsymbol{f}_{i}^{n}\right)-\varphi_1\left(\boldsymbol{f}_{i, j}^{n}\right)\right)\right\}_{j=1}^{s_l}, \nonumber
\end{equation}
where $\boldsymbol{f}_{i, j}^{n}$ is the neighbor feature of the feature $\boldsymbol{f}_{i}^{n}$, $\varphi_1$ is the MLP layer, $\phi_1$ is the skip-connection block, and $s_l$ represents the scale of $k$-NN with $l=1, 2$ in default. To achieve scale adaptation, we use an \emph{Attentional Feature Fusion} architecture to aggregate the features, thus benefiting from both the small and large scales. The fusion process is formulated as
\begin{equation}
\fontsize{9pt}{\baselineskip}\selectfont
M\left(\boldsymbol{f}_{i}^{s_1}, \boldsymbol{f}_{i}^{s_2}\right)=\mathrm{sigmoid}\left(\varphi_2\left(\mathrm{MaxPool}\left\{\boldsymbol{f}_{i}^{s_1}+\boldsymbol{f}_{i}^{s_2}\right\}_{i=1}^{N}\right)\right), \nonumber
\end{equation}
\begin{equation}
\fontsize{9pt}{\baselineskip}\selectfont
\boldsymbol{f}_{i}=\varphi_3\left(\boldsymbol{f}_{i}^{s_1}\cdot M\left(\boldsymbol{f}_{i}^{s_1}, \boldsymbol{f}_{i}^{s_2}\right)+\boldsymbol{f}_{i}^{s_2}\cdot \left(1-M\left(\boldsymbol{f}_{i}^{s_1}, \boldsymbol{f}_{i}^{s_2}\right)\right)\right), \nonumber
\end{equation}
where $\boldsymbol{f}_{i}^{s_1}$ abd $\boldsymbol{f}_{i}^{s_2}$ are the local structures with different scales of the feature $\boldsymbol{f}_{i}$, $\varphi_2$ and $\varphi_3$ are the MLP layers, and $N$ represents the cardinality of the input point cloud patch.

\subsubsection{Hierarchical Geometric Information Fusion}
To leverage various point set sizes effectively, it is common to employ a downsampling strategy for multi-scale global feature extraction~\cite{qi2017pointnet++,zhu2021adafit,li2022hsurf, qin2022geometric}. While this approach enriches the global geometric representation, there is a risk of losing the large-scale global information and local structures extracted during forward propagation. To mitigate this issue, as illustrated in Fig.~\ref{fig: hgif}, we introduce a \emph{hierarchical architecture} that integrates cross-scale global features and local structures during the downsampling of input point sets, \ie, 
\begin{equation}
\fontsize{9pt}{\baselineskip}\selectfont
\boldsymbol{f}_{i}^{N_{h+1}}=\varphi_4\left(\boldsymbol{G}_{N_h}, \boldsymbol{G}_{N_{h-1}}, \boldsymbol{g}_{i}^{N_{h+1}}\right) + \boldsymbol{f}_{i}^{N_h}, i=1, ..., N_{h+1}, \nonumber
\end{equation}
where $\varphi_4$ is the MLP layer, $\boldsymbol{G}_{N_h}$ represents the global feature of scale $N_h$, $\boldsymbol{g}_{i}^{N_{h+1}}$ represents the local structure, and $N_{h+1}\leq N_{h}\leq N_{h-1}$. During the Hierarchical Geometric Information Fusion, $\boldsymbol{G}_{N_h}$ is formulated as
\begin{equation}
\boldsymbol{G}_{N_h}=\varphi_6\left(\mathrm{MaxPool}\left\{\varphi_5\left(\boldsymbol{f}_{i}^{N_h}\right)\right\}_{i=1}^{N_h}\right), \nonumber
\end{equation}
where $\varphi_5$ and $\varphi_6$ are MLP layers. Meanwhile, the local structure $\boldsymbol{g}_{i}^{N_{h+1}}$ is captured by
\begin{small}
\begin{equation}
\boldsymbol{g}_{i}^{N_{h+1}}=\mathrm{MaxPool}\left\{\varphi_7\left(\boldsymbol{g}_{i, j}^{N_{h}}\right)\right\}_{j=1}^{s}+\boldsymbol{g}_{i}^{N_{h}}, i=1, ..., N_{h+1}, \nonumber
\end{equation}
\end{small}where $\boldsymbol{g}_{i, j}^{N_{h}}$ is the neighbor feature of point $\boldsymbol{p}_i$ in the scope of the scale $N_{h+1}$, $s$ is the scale of the neighbor features, and $\varphi_7$ represents the corresponding MLP layer. Moreover,  recognizing the pivotal role of positional information in normal estimation, we aim to prevent the loss of such information and capture the local structures within both geometric and high-level spatial contexts in the Hierarchical Geometric Information Fusion module. To achieve this, we utilize a range of local features across the \emph{odd} and \emph{even} hierarchical layers. The local features in the odd hierarchical layers are characterized by
\begin{equation}
\label{equ: local_feature1}
\boldsymbol{g}_{i, j}^{o}=\text{Concat}\left(\boldsymbol{p}_{i}, \boldsymbol{p}_{i}-\boldsymbol{p}_{i, j}, \varphi_8\left(\boldsymbol{p}_{i}-\boldsymbol{p}_{i, j}\right)\right), \nonumber
\end{equation}
where $\boldsymbol{p}_{i, j}$ is the neighbor coordinate of the point $\boldsymbol{p}_{i}$ and $\varphi_8$ represents the MLP layer. Simultaneously, the even layers concentrate more on high-level features, defined as 
\begin{equation}
\label{equ: local_feature2}
\boldsymbol{g}_{i, j}^{e}=\text{Concat}\left(\boldsymbol{p}_{i}, \boldsymbol{p}_{i}-\boldsymbol{p}_{i, j}, \boldsymbol{f}_{i}-\boldsymbol{f}_{i, j}\right), \nonumber
\end{equation}
where $\boldsymbol{f}_{i}$ and $\boldsymbol{f}_{i, j}$ are the high-level features of $\boldsymbol{p}_{i}$ and its neighbor, respectively.

\subsubsection{Unoriented Normal Estimation}
\begin{figure}[t]
\centering
\includegraphics[width=\linewidth]{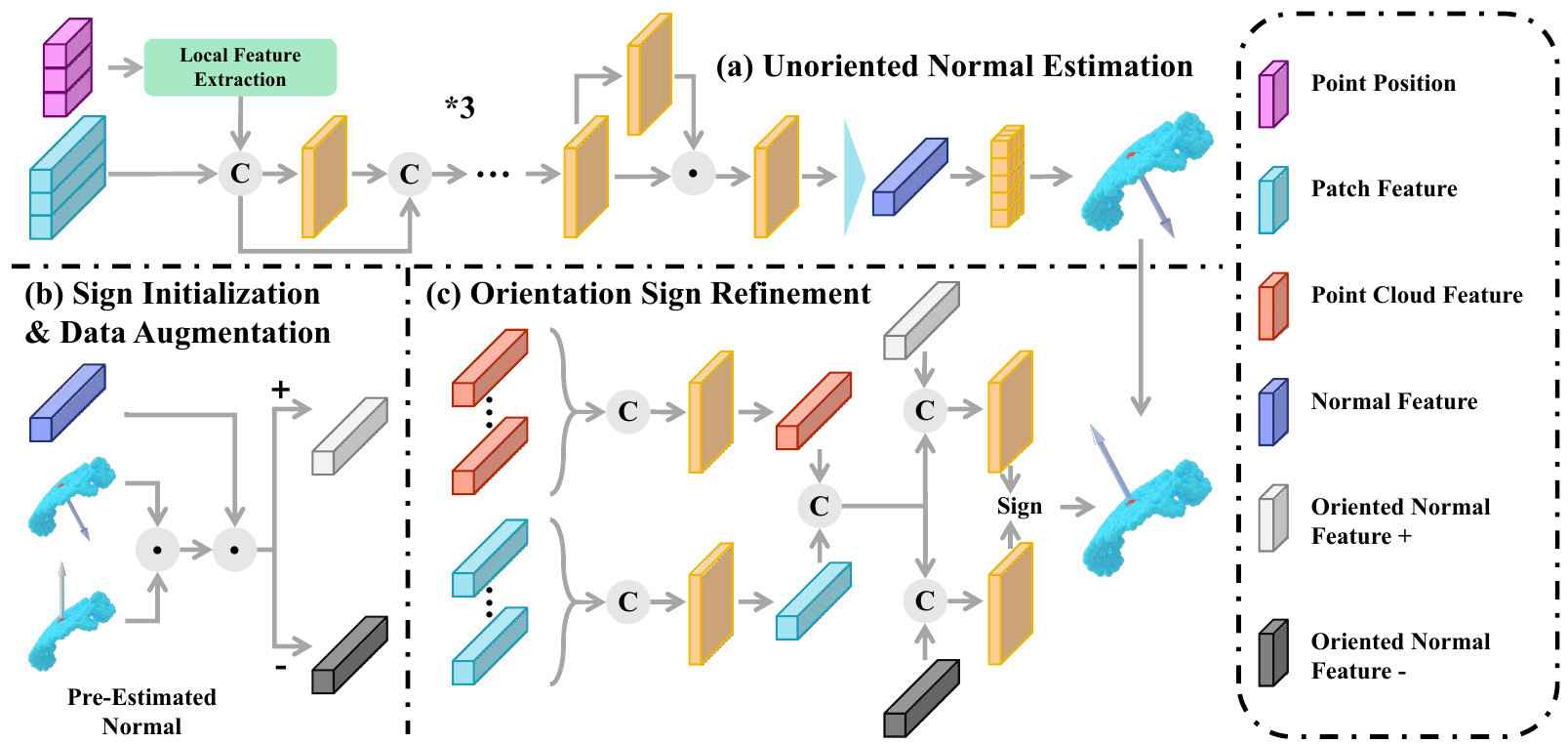}
\vskip -0.2cm
\caption{The orientated normal refinement module with the data augmentation in the feature space.}
\label{fig: refine}
\vskip -0.3cm
\end{figure}

Within the Refinement Module, we incorporate two branches for orientated normal estimation, \ie, unoriented normal estimation and orientation sign refinement. Point coordinates serve as fundamental attributes in point cloud processing, and the spatial interrelations, like distances between points, can significantly influence the inference process of point cloud neural networks. To explore this concept further, we introduce \emph{Position Feature Fusion} and \emph{Weighted Normal Prediction} into the unoriented normal estimation part. As shown in Fig.~\ref{fig: refine}(a), in the Position Feature Fusion step, we embed the neighbor coordinates of each point and merge them with the extracted patch feature using skip connections. This process can be expressed as
\begin{equation}
\boldsymbol{f'}_{i}^{\boldsymbol{P}}=\phi_2\left(\boldsymbol{f}_{i}^{\boldsymbol{P}}, \Psi_\mathrm{LFE}^s\left(\boldsymbol{p}_{i}, \boldsymbol{p}_{i, j}\right)\right), \nonumber
\end{equation}
where $\boldsymbol{p}_{i, j}$ is the neighbor coordinate of the point $\boldsymbol{p}_{i}$, $\boldsymbol{f}_{i}^{\boldsymbol{P}}$ is the extracted patch feature of $\boldsymbol{p}_{i}$, $s$ represents the neighbor scale, $\Psi_\mathrm{LFE}^s$ is the Local Feature Extraction block and $\phi_2$ is the skip connection block. Subsequently, we predict weights based on the position information of each point and use the weighted pooling to acquire the normal feature. The unoriented normal vector of the query point can be predicted as
\begin{equation}
\hat{\boldsymbol{n}}_{\mathrm{u}}=\boldsymbol{F}_{\boldsymbol{n}}\boldsymbol{W}, \nonumber
\end{equation}where
\begin{equation*}
\begin{small}
\boldsymbol{F}_{\boldsymbol{n}}=\mathrm{MaxPool}\left\{\varphi_{10}\left(\boldsymbol{f'}_{i}^{\boldsymbol{P}}\cdot \mathrm{softmax}\left(\varphi_{9}\left(\boldsymbol{f'}_{i}^{\boldsymbol{P}}\right)\right)\right)\right\}_{i=1}^{M}, \nonumber
\end{small}
\end{equation*} is the normal feature and $\boldsymbol{W}\in\mathbb{R}^{dim\times3}$. $\varphi_9$ and $\varphi_{10}$ are the MLP layers.

\subsubsection{Orientation Sign Refinement}
In the orientation sign refinement branch, we leverage the global features of the input patch $\boldsymbol{P}$ and the raw point cloud subsample set $\boldsymbol{D}$ to enhance the initialized orientation sign:
\begin{equation}
{sign} = \varphi_{11}\left(\boldsymbol{F}_{\boldsymbol{n}}^{sign}, \boldsymbol{F}_{\boldsymbol{D}}, \boldsymbol{F}_{\boldsymbol{P}}\right), \nonumber
\end{equation} where the pre-oriented sign is mapped to the normal feature space $\boldsymbol{F}_{\boldsymbol{n}}^{{sign}}$. Specifically, the global descriptions of $\boldsymbol{P}$ and $\boldsymbol{D}$ are extracted based on the multi-scale global features in the Hierarchical Geometric Information Fusion
\begin{equation}
\boldsymbol{F}_{\boldsymbol{D}} = \varphi_{12}\left(\text{Concat}\left(\boldsymbol{G}_{\boldsymbol{D}}^{N_1}\cdots\boldsymbol{G}_{\boldsymbol{D}}^{N_a}\right)\right), \nonumber
\end{equation}
\begin{equation}
\boldsymbol{F}_{\boldsymbol{P}} = \varphi_{13}\left(\text{Concat}\left(\boldsymbol{G}_{\boldsymbol{P}}^{N_1}\cdots\boldsymbol{G}_{\boldsymbol{P}}^{N_b}\right)\right), \nonumber
\end{equation}
where $\boldsymbol{G}$ represents the global feature of $\boldsymbol{P}$ and $\boldsymbol{D}$ with a certain scale, $a$ and $b$ are the numbers of scales in the downsampling process of $\boldsymbol{P}$ and $\boldsymbol{D}$, respectively. Moreover, we find an efficient way to represent the initialized orientation sign. As the final unoriented normal is estimated by a single linear layer without bias, the sign of the vector in the feature space has an absolute correspondence with the sign of the output normal vector. Namely, the initialized sign can be projected to the normal feature $\boldsymbol{F}_{\boldsymbol{n}}$ in the unoriented normal estimation branch for orientation sign refinement
\begin{equation}
\boldsymbol{F}_{\boldsymbol{n}}^{sign+} = \boldsymbol{F}_{\boldsymbol{n}}\cdot{sgn}_\mathrm{MST}, \nonumber
\end{equation}
where ${sgn}_\mathrm{MST}={sgn}(\hat{\boldsymbol{n}}_\mathrm{u}\cdot\boldsymbol{n}_\mathrm{init})$. Moreover, due to the correspondence between the feature space and the estimated normal, the negative example can be generated by the inverse feature $\boldsymbol{F}_{\boldsymbol{n}}^{sign-}=-\boldsymbol{F}_{\boldsymbol{n}}^{sign+}$. Since the data augmentation is processed in the high-level feature space, the network can be trained to observe more diverse samples with slight additional time. Notably, when applying the data augmentation, directly employing one layer for sign refinement will force the ratio of positive to negative signs in the initialization to be $1:1$, which results in the loss of important prior information from the sign initialization. To avoid this problem, we adopt two MLP layers to refine the positive and negative initialized sign and the output sign is the one with a higher probability
\begin{equation}
\fontsize{8pt}{\baselineskip}\selectfont
sign = sgn\left(\varphi_{+}\left(\boldsymbol{F}_{\boldsymbol{n}}^{sign+}, \boldsymbol{F}_{\boldsymbol{D}}, \boldsymbol{F}_{\boldsymbol{P}}\right)-\varphi_{-}\left(\boldsymbol{F}_{\boldsymbol{n}}^{sign-}, \boldsymbol{F}_{\boldsymbol{D}}, \boldsymbol{F}_{\boldsymbol{P}}\right)\right). \nonumber
\end{equation}

\begin{table*}[htbp]
\centering
\small
\setlength{\tabcolsep}{4pt}
\renewcommand\arraystretch{1}
\caption{Quantitative comparisons of unoriented normal estimation in terms of RMSE and CND on the PCPNet dataset. \textbf{Bold} values indicate the best estimator.}
\begin{tabular}{l|cccc|cc|c|cccc|cc|c}
\hline
{\multirow{3}{*}{Method}}&\multicolumn{7}{c|}{{{RMSE}}}& \multicolumn{7}{c}{{CND}}  \\ \cline{2-15}
& \multicolumn{4}{c|}{{Noise ($\sigma$)}}& \multicolumn{2}{c|}{{Density}}& \multirow{2}{*}{{Ave.}}       & \multicolumn{4}{c|}{{Noise ($\sigma$)}}& \multicolumn{2}{c|}{{Density}} & \multirow{2}{*}{{Ave.}}  \\
&None          & 0.12\%        & 0.6\%          & \multicolumn{1}{c|}{1.2\%}          & Stripe       & \multicolumn{1}{c|}{Grad.}      & & None          & 0.12\%        & 0.6\%          & \multicolumn{1}{c|}{1.2\%}          & Stripe    & \multicolumn{1}{c|}{Grad.}      & \\ \hline
\hline
PCA~\cite{hoppe1992surface}                          & 12.28         & 12.86         & 18.40          & \multicolumn{1}{c|}{27.61}                         & 13.63         & \multicolumn{1}{c|}{12.79}                        & 16.26                     & 12.28         & 12.79         & 16.41          & \multicolumn{1}{c|}{24.46}                                  & 13.63         & \multicolumn{1}{c|}{12.79}                        & 15.39                     \\ 
n-jet~\cite{cazals2005estimating}                         & 12.32         & 12.82         & 18.34          & \multicolumn{1}{c|}{27.77}                         & 13.36         & \multicolumn{1}{c|}{13.09}                        & 16.29                     & 12.32         & 12.77         & 16.36          & \multicolumn{1}{c|}{24.67}                                  & 13.36         & \multicolumn{1}{c|}{13.09}                        & 15.43                     \\
PCPNet~\cite{guerrero2018pcpnet}                 & 7.22          & 10.94         & 18.18          & \multicolumn{1}{c|}{22.65}                         & 8.20          & \multicolumn{1}{c|}{8.06}                         & 12.54                     & 7.22          & 10.77         & 16.48          & \multicolumn{1}{c|}{19.28}                                  & 8.20          & \multicolumn{1}{c|}{8.06}                         & 11.69                     \\
Nesti-Net~\cite{ben2019nesti}                         & 7.09          & 10.90         & 18.02          & \multicolumn{1}{c|}{22.47}                         & 8.24          & \multicolumn{1}{c|}{8.45}                         & 12.53                     & 7.09          & 10.77         & 16.19          & \multicolumn{1}{c|}{18.53}                                  & 8.24          & \multicolumn{1}{c|}{8.45}                         & 11.55                     \\
DeepFit~\cite{ben2020deepfit}                         & 6.55          & 9.44          & 16.84          & \multicolumn{1}{c|}{23.46}                         & 7.72          & \multicolumn{1}{c|}{7.29}                         & 11.88                     & 6.55          & 9.24          & 14.09          & \multicolumn{1}{c|}{19.35}                                  & 7.72          & \multicolumn{1}{c|}{7.29}                         & 10.71                     \\
AdaFit~\cite{zhu2021adafit}                       & 4.64          & 9.16          & 16.54          & \multicolumn{1}{c|}{22.11}                         & 5.22          & \multicolumn{1}{c|}{5.24}                         & 10.49                     & 4.64          & 8.89          & 13.75          & \multicolumn{1}{c|}{17.51}                                  & 5.22          & \multicolumn{1}{c|}{5.24}                         & 9.21                      \\
GraphFit~\cite{li2022graphfit}                       & 3.92          & 8.73          & 16.21          & \multicolumn{1}{c|}{21.90}                         & 4.60          & \multicolumn{1}{c|}{4.52}                         & 9.98                      & 3.92          & 8.46          & 13.22          & \multicolumn{1}{c|}{17.41}                                  & 4.60          & \multicolumn{1}{c|}{4.52}                         & 8.69                      \\
HSurf-Net~\cite{li2022hsurf}                   & 3.91          & 8.82          & 16.22          & \multicolumn{1}{c|}{21.68}                         & 4.56          & \multicolumn{1}{c|}{4.55}                         & 9.96                      & 3.91          & 8.54          & 13.23          & \multicolumn{1}{c|}{16.85}                                  & 4.56          & \multicolumn{1}{c|}{4.55}                         & 8.61                      \\
Du~\etal~\cite{du2023rethinking}                       & 3.78          & 8.77          & 16.13          & \multicolumn{1}{c|}{21.70}                         & 4.50          & \multicolumn{1}{c|}{4.61}                         & 9.92                      & 3.78          & 8.49          & 13.12          & \multicolumn{1}{c|}{17.13}                                  & 4.50          & \multicolumn{1}{c|}{4.61}                         & 8.61                      \\
SHS-Net~\cite{li2023shs}                      & 3.82          & 8.68          & 16.41          & \multicolumn{1}{c|}{21.79}                         & 4.53          & \multicolumn{1}{c|}{4.49}                         & 9.95                      & 3.82          & 8.43          & 13.49          & \multicolumn{1}{c|}{17.25}                                  & 4.53          & \multicolumn{1}{c|}{4.49}                         & 8.67                      \\
MSECNet~\cite{xiu2023msecnet}                       & 3.64          & 8.85          & 16.26          & \multicolumn{1}{c|}{\textbf{21.24}}                & 4.21          & \multicolumn{1}{c|}{\textbf{4.22}}                & 9.74                      & 3.64          & 8.51          & 13.85          & \multicolumn{1}{c|}{16.75}                                  & 4.21          & \multicolumn{1}{c|}{\textbf{4.22}}                & 8.53                      \\
NGLO~\cite{li2023neural}                       & 3.63          & 8.83          & 16.19          & \multicolumn{1}{c|}{21.59}                         & 4.40          & \multicolumn{1}{c|}{4.23}                         & 9.81                      & 3.63          & 8.59          & 13.16          & \multicolumn{1}{c|}{16.69}                                  & 4.40          & \multicolumn{1}{c|}{4.23}                         & 8.45                      \\
NeuralGF~\cite{li2024neuralgf}                      & 7.81          & 9.62          & 18.64          & \multicolumn{1}{c|}{25.14}                         & 8.71          & \multicolumn{1}{c|}{8.75}                         & 13.11                     & 7.81          & 9.31          & 15.93          & \multicolumn{1}{c|}{20.84}                                  & 8.71          & \multicolumn{1}{c|}{8.75}                         & 11.89                     \\
\cellcolor{boxbody}{Ours}                                            & \cellcolor{boxbody}{\textbf{3.48}} & \cellcolor{boxbody}{\textbf{8.57}} & \cellcolor{boxbody}{\textbf{16.05}} & \cellcolor{boxbody}{21.76} & \cellcolor{boxbody}{\textbf{4.16}} & \cellcolor{boxbody}{4.25} & \cellcolor{boxbody}{\textbf{9.71}}          & \cellcolor{boxbody}{\textbf{3.48}} & \cellcolor{boxbody}{\textbf{8.30}} & \cellcolor{boxbody}{\textbf{12.40}} & \cellcolor{boxbody}{\textbf{16.10}} & \cellcolor{boxbody}{\textbf{4.16}} & \cellcolor{boxbody}{4.25} & \cellcolor{boxbody}{\textbf{8.12}}             \\ \hline
\end{tabular}
\label{tab: pcpnet_u}
\end{table*}
\small
\setlength{\tabcolsep}{4pt}

\begin{table*}[htbp]
\centering
\small
\setlength{\tabcolsep}{4pt}
\renewcommand\arraystretch{1}
\caption{Comparisons of unoriented normal estimation regarding RMSE and CND on the FamousShape dataset.}
\begin{tabular}{l|cccc|cc|c|cccc|cc|c}
\hline
{\multirow{3}{*}{Method}}&\multicolumn{7}{c|}{{{RMSE}}}& \multicolumn{7}{c}{{CND}}  \\ \cline{2-15}
& \multicolumn{4}{c|}{{Noise ($\sigma$)}}& \multicolumn{2}{c|}{{Density}}& \multirow{2}{*}{{Ave.}}       & \multicolumn{4}{c|}{{Noise ($\sigma$)}}& \multicolumn{2}{c|}{{Density}} & \multirow{2}{*}{{Ave.}}  \\
&None          & 0.12\%        & 0.6\%          & \multicolumn{1}{c|}{1.2\%}          & Stripe       & \multicolumn{1}{c|}{Grad.}      & & None          & 0.12\%        & 0.6\%          & \multicolumn{1}{c|}{1.2\%}          & Stripe    & \multicolumn{1}{c|}{Grad.}      & \\ \hline
\hline
PCA~\cite{hoppe1992surface}                & 19.85         & 20.56          & 31.34          & \multicolumn{1}{c|}{45.06}                                  & 19.80         & \multicolumn{1}{c|}{18.50}                                 & 25.85                     & 19.85         & 20.47          & 27.57          & \multicolumn{1}{c|}{40.89}                                  & 19.80         & \multicolumn{1}{c|}{18.50}                                 & 24.51                     \\
n-jet~\cite{cazals2005estimating}                & 20.05         & 20.53          & 31.36          & \multicolumn{1}{c|}{45.25}                                  & 18.83         & \multicolumn{1}{c|}{18.64}                                 & 25.78                     & 20.05         & 20.46          & 27.60          & \multicolumn{1}{c|}{41.20}                                  & 18.83         & \multicolumn{1}{c|}{18.64}                                 & 24.46                     \\
PCPNet~\cite{guerrero2018pcpnet}                 & 11.66         & 17.85          & 31.91          & \multicolumn{1}{c|}{40.18}                                  & 11.54         & \multicolumn{1}{c|}{11.12}                                 & 20.71                     & 11.66         & 17.52          & 29.39          & \multicolumn{1}{c|}{35.23}                                  & 11.54         & \multicolumn{1}{c|}{11.12}                                 & 19.41                     \\
Nesti-Net~\cite{ben2019nesti}            & 11.36         & 17.37          & 31.72          & \multicolumn{1}{c|}{39.52}                                  & 11.71         & \multicolumn{1}{c|}{11.55}                                 & 20.54                     & 11.36         & 16.99          & 29.17          & \multicolumn{1}{c|}{33.57}                                  & 11.71         & \multicolumn{1}{c|}{11.55}                                 & 19.06                     \\
DeepFit~\cite{ben2020deepfit}              & 11.44         & 16.82          & 29.87          & \multicolumn{1}{c|}{40.05}                                  & 11.68         & \multicolumn{1}{c|}{10.72}                                 & 20.10                     & 11.44         & 16.43          & 25.65          & \multicolumn{1}{c|}{33.33}                                  & 11.68         & \multicolumn{1}{c|}{10.72}                                 & 18.21                     \\
AdaFit~\cite{zhu2021adafit}                 & 8.27          & 15.90          & 29.89          & \multicolumn{1}{c|}{38.75}                                  & 7.93          & \multicolumn{1}{c|}{7.92}                                  & 18.11                     & 8.27          & 15.47          & 26.11          & \multicolumn{1}{c|}{32.09}                                  & 7.93          & \multicolumn{1}{c|}{7.92}                                  & 16.30                     \\
GraphFit~\cite{li2022graphfit}               & 7.15          & 15.23          & 29.57          & \multicolumn{1}{c|}{38.82}                                  & 7.06          & \multicolumn{1}{c|}{6.91}                                  & 17.46                     & 7.15          & 14.67          & 25.56          & \multicolumn{1}{c|}{32.49}                                  & 7.06          & \multicolumn{1}{c|}{6.91}                                  & 15.64                     \\
HSurf-Net~\cite{li2022hsurf}                   & 7.22          & 15.62          & 29.49          & \multicolumn{1}{c|}{38.60}                                  & 7.06          & \multicolumn{1}{c|}{6.98}                                  & 17.50                     & 7.22          & 15.15          & 25.43          & \multicolumn{1}{c|}{32.17}                                  & 7.06          & \multicolumn{1}{c|}{6.98}                                  & 15.67                     \\
Du~\etal~\cite{du2023rethinking}                      & 7.06          & 15.29          & 29.43          & \multicolumn{1}{c|}{38.68}                                  & 6.98          & \multicolumn{1}{c|}{6.84}                                  & 17.38                     & 7.06          & 14.76          & 25.32          & \multicolumn{1}{c|}{32.20}                                  & 6.98          & \multicolumn{1}{c|}{6.84}                                  & 15.53                     \\
SHS-Net~\cite{li2023shs}                         & 7.09          & 15.46          & 29.63          & \multicolumn{1}{c|}{38.65}                                  & 7.12          & \multicolumn{1}{c|}{6.98}                                  & 17.49                     & 7.09          & 14.95          & 25.76          & \multicolumn{1}{c|}{32.68}                                  & 7.12          & \multicolumn{1}{c|}{6.98}                                  & 15.76                     \\
MSECNet~\cite{xiu2023msecnet}                    & \textbf{6.62} & 15.41          & 29.28          & \multicolumn{1}{c|}{38.78}                                  & \textbf{6.67} & \multicolumn{1}{c|}{\textbf{6.65}}                         & 17.24                     & \textbf{6.62} & 14.94          & 25.61          & \multicolumn{1}{c|}{31.75}                                  & \textbf{6.67} & \multicolumn{1}{c|}{\textbf{6.65}}                         & 15.37                     \\
NGLO~\cite{li2023neural}                     & 6.89          & 15.51          & 29.46          & \multicolumn{1}{c|}{38.79}                                  & 6.93          & \multicolumn{1}{c|}{6.86}                                  & 17.41                     & 6.89          & 14.97          & 25.29          & \multicolumn{1}{c|}{32.55}                                  & 6.93          & \multicolumn{1}{c|}{6.86}                                  & 15.58                     \\
NeuralGF~\cite{li2024neuralgf}                  & 12.34         & 15.80          & 31.41          & \multicolumn{1}{c|}{40.89}                                  & 13.33         & \multicolumn{1}{c|}{13.12}                                 & 21.15                     & 12.34         & 15.22          & 26.93          & \multicolumn{1}{c|}{33.96}                                  & 13.33         & \multicolumn{1}{c|}{13.12}                                 & 19.15                     \\
\cellcolor{boxbody}{Ours}                                            & \cellcolor{boxbody}{6.88} & \cellcolor{boxbody}{\textbf{14.91}} & \cellcolor{boxbody}{\textbf{29.02}} & \cellcolor{boxbody}{\textbf{38.71}} & \cellcolor{boxbody}{6.80} & \cellcolor{boxbody}{\textbf{6.65}} & \cellcolor{boxbody}{\textbf{17.16}}          & \cellcolor{boxbody}{6.88} & \cellcolor{boxbody}{\textbf{14.32}} & \cellcolor{boxbody}{\textbf{23.94}} & \cellcolor{boxbody}{\textbf{30.76}} & \cellcolor{boxbody}{6.80} & \cellcolor{boxbody}{\textbf{6.65}} & \cellcolor{boxbody}{\textbf{14.89}}             \\ \hline
\end{tabular}
\label{tab: famousshape_u}
\vskip -0.3cm
\end{table*}
\small
\setlength{\tabcolsep}{4pt}

\begin{figure*}[h]
\centering
\includegraphics[width=0.9\linewidth]{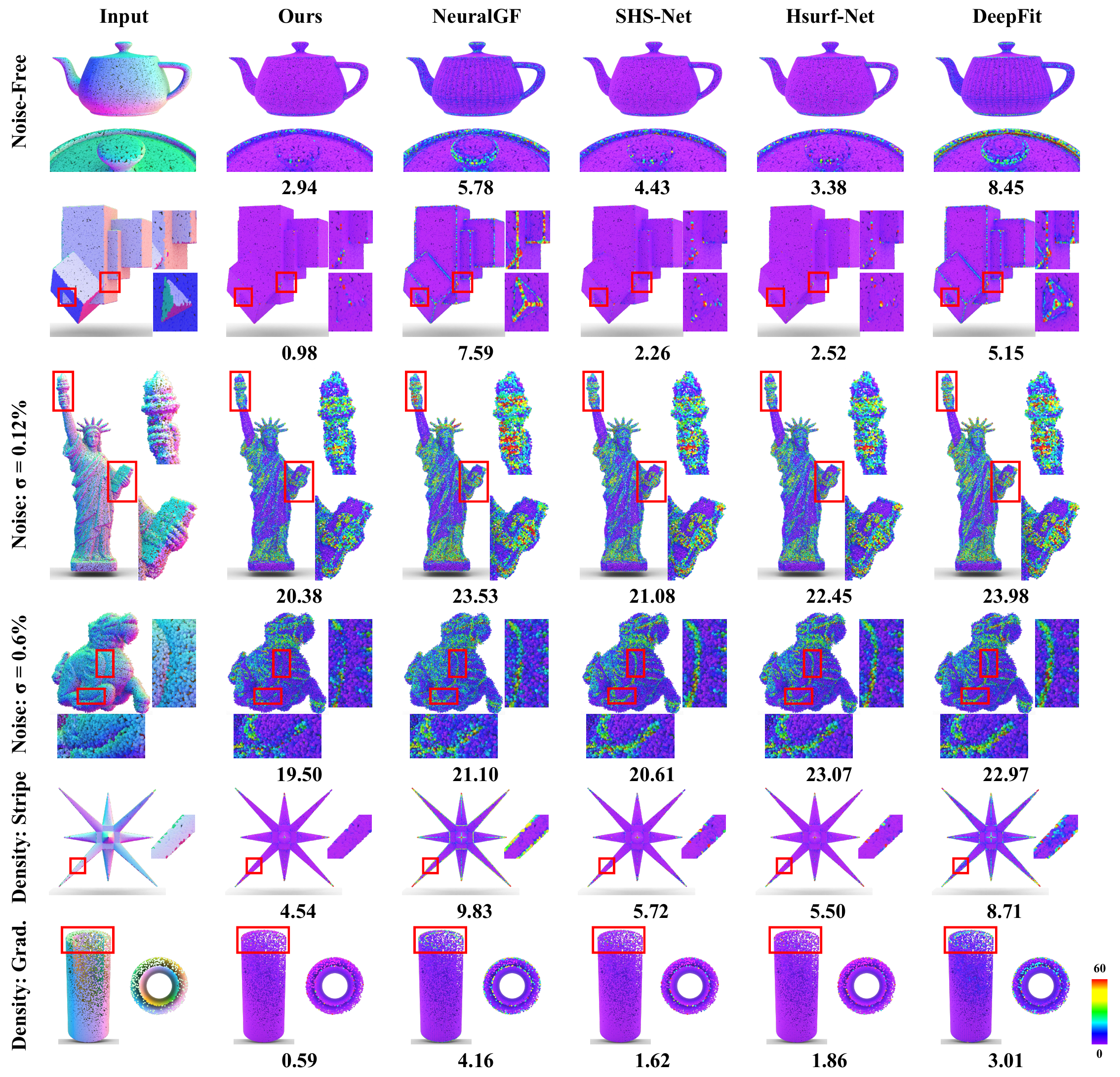}
\caption{Qualitative comparisons of unoriented  normal estimation on the PCPNet and FamousShape datasets. We use the heat map to visualize the CND error.}
\label{fig: pcp_u}
\end{figure*}

\subsubsection{Loss Function}
To reconcile the annotated normal with the geometric fluctuations induced by noise in the neighbor of the query point, we redefine the sine loss using the  CND metric. Specifically, we designate the normal $\boldsymbol{n}_{\tilde{\boldsymbol{{p}}}}$ of the nearest neighbor point $\tilde{\boldsymbol{{p}}}$ in the corresponding noise-free point cloud $\tilde{\mathcal{P}}$ as the ground-truth
\begin{equation}
\mathcal{L}_1=\left\Vert\boldsymbol{n}_{\tilde{\boldsymbol{p}}}\times\hat{\boldsymbol{n}}_{\boldsymbol{p}}\right\Vert. \nonumber
\end{equation}Meanwhile, we use the $\boldsymbol{z}$-direction transformation loss to constrain the output rotation matrix $\boldsymbol{\mathbf{R}_\text{QSTN}}\in\mathbb{R}^{3\times 3}$ of the QSTN~\cite{du2023rethinking}
\begin{equation}
\mathcal{L}_2=\left\Vert\boldsymbol{n}_{\tilde{\boldsymbol{p}}}\boldsymbol{\mathbf{R}_\text{QSTN}}\times{\boldsymbol{z}}\right\Vert, \nonumber
\end{equation}
where $\boldsymbol{z}=\left(0, 0, 1\right)$. Additionally, to make full use of the spatial relationships between data points, we adopt the weighted loss similar to Zhang~\etal~\cite{zhang2022geometry}
\begin{equation}
\mathcal{L}_3=\frac{1}{M}\sum_{i=1}^{M}(w_i-\hat{w}_i)^2, \nonumber
\end{equation}
where $\hat{w}$ are the predicted weights for each data point, $M$ represents the cardinality of the downsampled patch, $w_i=\exp(-\left(\boldsymbol{p}_i\cdot\boldsymbol{n}_{\tilde{\boldsymbol{p}}}\right)^2/\delta^2)$ and $\delta=\max\left(0.05^2,0.3\sum_{i=1}^M\left(\boldsymbol{p}_i\cdot\boldsymbol{n}_{\tilde{\boldsymbol{p}}}\right)^2/M\right)$, where $\boldsymbol{p}_i$ is the point in the downsampled patch. 

For the orientation sign refinement, we adopt the \emph{binary cross entropy} $H$~\cite{erler2020points2surf} to define the sign classification loss
\begin{equation}
\mathcal{L}_4^+=H\Big(\sigma\big({sgn}^+\big), [{sgn}_\text{MST}\cdot{sgn}_\text{GT}]\Big), \nonumber
\end{equation}
\begin{equation}
\mathcal{L}_4^-=H\Big(\sigma\big({sgn}^-\big), [-{sgn}_\text{MST}\cdot{sgn}_\text{GT}]\Big), \nonumber
\end{equation}
\begin{equation}
\mathcal{L}_4=\mathcal{L}_4^++\mathcal{L}_4^-, \nonumber
\end{equation}
where $\sigma$ is a logistic function that converts the sign logits ${sgn}^+$ and ${sgn}^-$ estimated by $\varphi_{+}$ and $\varphi_{-}$ to probabilities. $[{sgn}_\text{MST}\cdot{sgn}_\text{GT}]$ is 1 if the initialized sign ${sgn}_\text{MST}$ is consist with the ground-truth orientation sign ${sgn}_\text{GT}$ and 0 otherwise. ${sgn}_\text{GT}$ is the orientation sign of the ground-truth normal $\boldsymbol{n}_{\tilde{
\boldsymbol{p}}}$ in the CND-modified loss $\mathcal{L}_1$. Moreover, we employ a \emph{contrastive loss} to make full use of the positive and negative examples
\begin{equation}
\mathcal{L}_5=\exp\left(-{\left(\sigma\left({sgn}^+\right)-\sigma\left({sgn}^-\right)\right)}^2\right), \nonumber
\end{equation}
Therefore, our final loss function is formally defined as
\begin{equation}
\mathcal{L}=\lambda_1\mathcal{L}_1+\lambda_2\mathcal{L}_2+\lambda_3\mathcal{L}_3+\lambda_4\mathcal{L}_4+\lambda_5\mathcal{L}_5, \nonumber
\end{equation}
where $\lambda_1=0.1$, $\lambda_2=0.5$, $\lambda_3=1$, $\lambda_4=0.1$, and $\lambda_5=0.1$ are weighting factors.

\section{Experimental Results}
In this section, we perform comprehensive experiments to validate the efficacy of the proposed method across diverse datasets, encompassing synthetic, real-world, indoor, and outdoor scenes.



\subsection{Implementation Details}
For the input patch, we fix the size at $N_P=700$ and utilize downsampling factors of $\rho_{\boldsymbol{P}}=\{2/3, 2/3, 2/3, 1\}$. The $k$-NN scales in the Multi-scale Local Feature extraction are set to $16$ and $32$, while $s=\{32, 32, 16, 16\}$ in the Hierarchical Geometric Information Fusion. For the subsample of the point cloud, we define the size as $N_D=1,200$, downsampling factors as $\rho_{\boldsymbol{D}}=\{1/2, 1/2, 1\}$, and the Local Feature Extraction size as $8$. During the Position Feature Fusion, the number of neighbor points is designated as 16. We adopt the AdamW optimizer~\cite{loshchilov2017decoupled} with an initial learning rate of $5 \times 10^{-4}$ for training. The learning rate is decayed following a cosine function. Our model is trained with a batch size of 64 on an NVIDIA A100 GPU over 900 epochs. We present more implementation details in the \emph{Supplementary Material}.


\subsection{Datasets}
We first adopt the synthetic dataset PCPNet~\cite{guerrero2018pcpnet} for comparison, which contains $8$ shapes for training and $18$ shapes for testing. All shapes are given as triangle meshes and densely sampled with 100K points. The sampled point clouds are augmented by introducing noise with a standard deviation of $0.12\%$, $0.6\%$, and $1.2\%$ with respect to the bounding box size of the model. Additionally, non-uniform samplings encompassing stripes and gradients are included. We also evaluate our method on the more challenging FamousShape dataset~\cite{li2023shs,li2024learning}, which features complex geometric structures for a deeper comparison.
To demonstrate  the generalization capability of our method, we further evaluate and compare the models trained on PCPNet using the real-world indoor SceneNN dataset~\cite{hua-pointwise-cvpr18} and the outdoor Semantic3D dataset~\cite{hackel2017isprs}. Moreover, to validate the versatility of our method in reconstruction tasks, we conduct additional comparisons on datasets with complex topology~\cite{xu2023globally} and wireframe point clouds~\cite{huang2019variational}.



\subsection{Evaluation Metrics}
To make thorough comparison, we adopt the proposed CND metric to assess the normal estimation results and compare it with the RMSE. The normal angle errors within RMSE and CND metrics are constrained between $0\degree$ and $90\degree$ for unoriented normal evaluation and between $0\degree$ and $180\degree$ for oriented normal evaluation. Additionally, the error distribution is visualized through the \emph{Area Under the Curve} (AUC), which can be found in the \emph{Supplementary Material}, along with more evaluation details.


\begin{figure*}[t]
\centering
\includegraphics[width=0.9\linewidth]{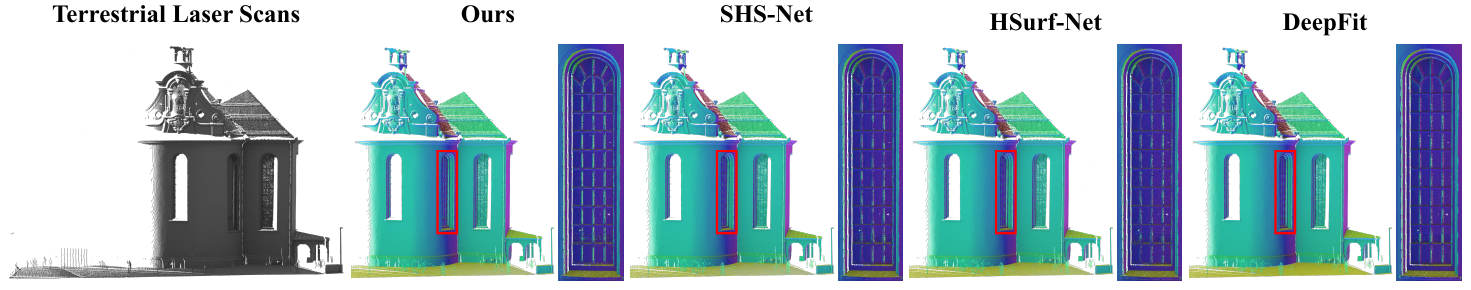}
\caption{Qualitative comparisons of unoriented normal estimation on the real-world Semantic3D dataset. Point clouds are colored by the estimated normals.}
\label{fig: semantic_u}
\vskip -0.3cm
\end{figure*}

\subsection{Unoriented Normal Comparison}

\subsubsection{Synthetic Dataset}
We first perform evaluation on unoriented normal estimation, where both traditional methods PCA~\cite{hoppe1992surface}, n-jet~\cite{cazals2005estimating}, and the latest learning-based methods PCPNet~\cite{guerrero2018pcpnet}, Nesti-Net~\cite{ben2019nesti}, DeepFit~\cite{ben2020deepfit}, AdaFit~\cite{zhu2021adafit}, GraphFit~\cite{li2022graphfit}, HSurf-Net~\cite{li2022hsurf}, Du~\etal~\cite{du2023rethinking}, SHS-Net~\cite{li2023shs}, MSECNet~\cite{xiu2023msecnet}, NGLO~\cite{li2023neural}, and NeuralGF~\cite{li2024neuralgf} are taken as baselines. All learning-based methods are trained using the same dataset. The statistical results of all compared approaches on the PCPNet and FamousShape datasets, assessed in terms of both RMSE and CND metrics, are summarized in Table~\ref{tab: pcpnet_u} and Table~\ref{tab: famousshape_u}. As observed, our method achieves the overall highest normal estimation accuracy across different scenarios, particularly in scenarios with noise. The CND metric, compared to RMSE, provides more accurate and reliable prediction assessments while addressing annotation discrepancies. Qualitative comparison results are presented in Fig.~\ref{fig: pcp_u}. Notably, our method exhibits the smallest errors in sharp areas and regions characterized by noise and intricate geometries. More qualitative results are presented in the \emph{Supplementary Material}.

\subsubsection{Real-world Dataset}
Next, we investigate the generalization capability of our method using the real-world indoor SceneNN dataset and outdoor Semantic3D~\cite{hackel2017isprs} dataset. Due to the ambiguity in judging the internal or external orientation of normals in point clouds not sampled from closed surfaces, we focus on reporting results for unoriented normals for these datasets. The results presented in Table~\ref{tab: scenennn} suggest that our method has the highest normal estimation accuracy on average across the SceneNN dataset. The qualitative results for the outdoor scenes,  presented in Fig.~\ref{fig: semantic_u}, also underscore our method's superiority. It is noticeable that our method successfully preserves more geometric details, such as the window grids in Fig.~\ref{fig: semantic_u}. We represent more qualitative results in the \emph{Supplementary Material}.



\subsection{Oriented Normal Comparison}
\begin{figure*}[t]
\centering
\includegraphics[width=0.9\linewidth]{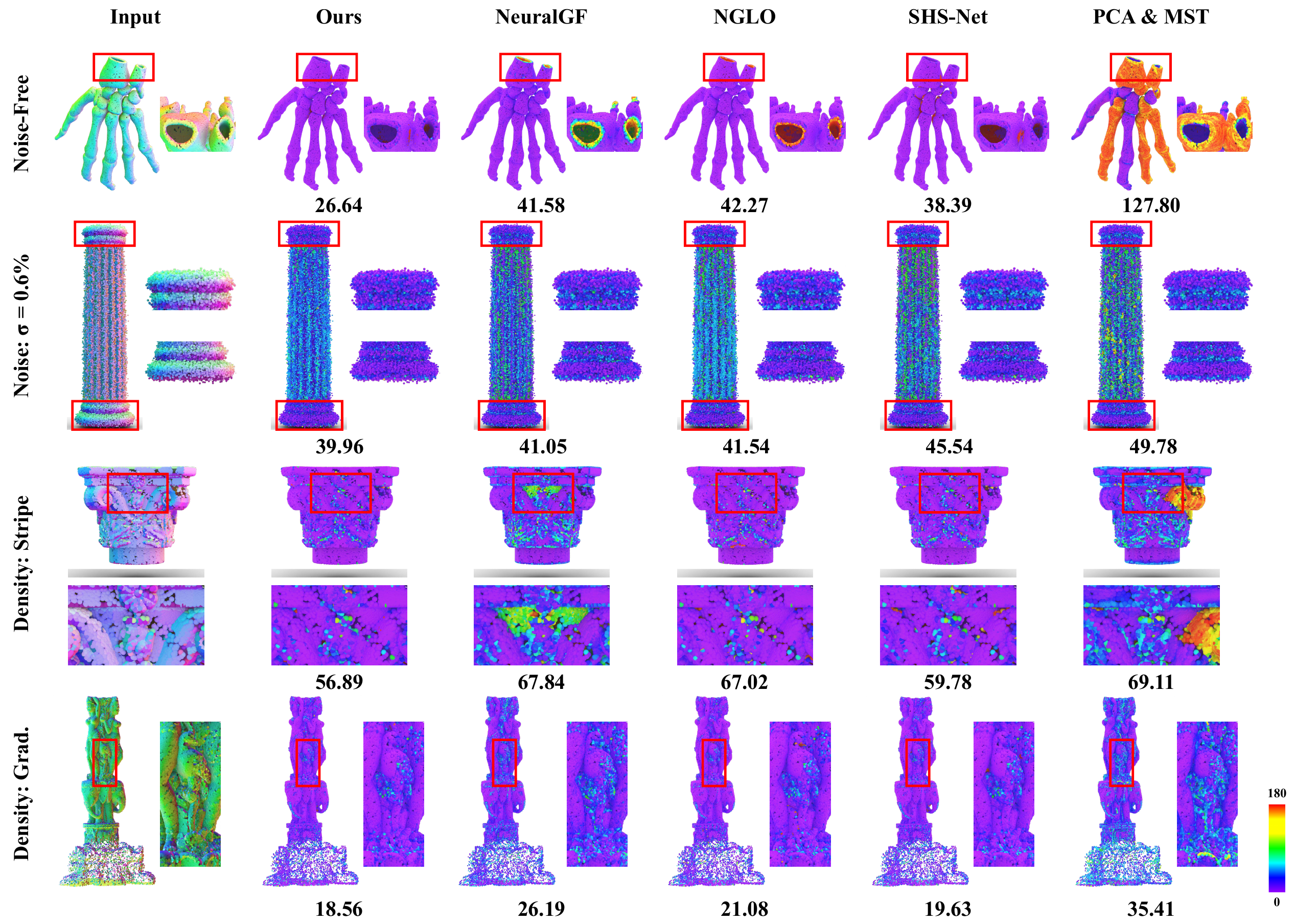}
\vskip -0.2cm
\caption{Qualitative comparisons of oriented normal estimation on the PCPNet and FamousShape datasets.}
\label{fig: vis_o}
\vskip -0.3cm
\end{figure*}

\begin{figure*}[!h]
\centering
\includegraphics[width=0.9\linewidth]{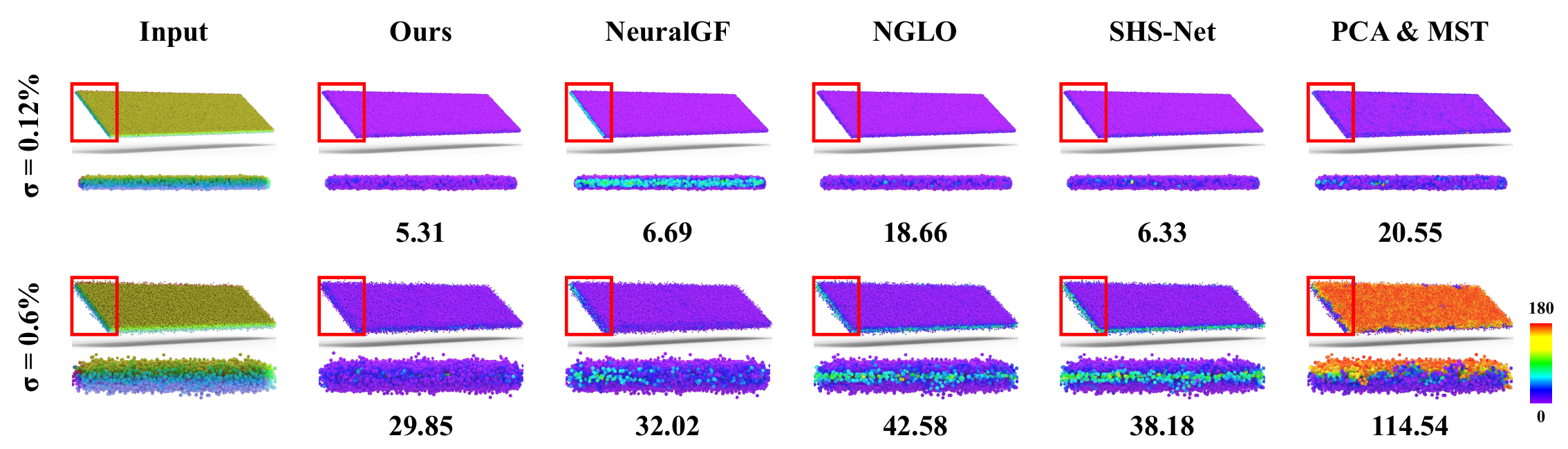}
\vskip -0.2cm
\caption{Qualitative comparisons of oriented normal estimation on the thin sheet structures with different noise levels.}
\label{fig: vis_sheet}
\vskip -0.3cm
\end{figure*}

For the comparison of oriented normal estimation, we select three representative unoriented normal estimation methods—PCA~\cite{hoppe1992surface}, AdaFit~\cite{zhu2021adafit}, and HSurf-Net~\cite{li2022hsurf}—along with three normal orientation methods—MST~\cite{hoppe1992surface}, SNO~\cite{schertler2017towards}, and ODP~\cite{metzer2021orienting}. These are combined in various configurations to create two-stage pipelines for estimating oriented normals. Additionally, we consider the latest learning-based approaches, such as PCPNet~\cite{guerrero2018pcpnet}, SHS-Net~\cite{li2023shs}, NGLO~\cite{li2023neural}, and NeuralGF~\cite{li2024neuralgf}, as baselines. In Tables~\ref{tab: pcpnet_o} and~\ref{tab: famous_o}, we report quantitative comparison results on the PCPNet and FamousShape datasets. Notably, our method achieves the most accurate results across almost all noise levels on the PCPNet dataset and delivers the best average outcomes on both PCPNet and FamousShape datasets.
Qualitative results in Fig.~\ref{fig: vis_o} demonstrate that our method can achieve accurate sign orientation results even in details like holes and small gaps, as well as providing more accurate normals in areas with complex geometries. Moreover, as shown in Fig.~\ref{fig: vis_sheet}, benefiting from the CND-modified loss and comprehensive geometric representation, our method successfully retains the detailed sharp edges of the challenging sheet structures to a significant degree across diverse noise settings.

\subsubsection{Refinement on Other Methods}
Subsequently, we demonstrate the adaptability of our OCMG-Net with respect to refining normals estimated by other approaches. As summarized in Table~\ref{tab: refine}, our method effectively enhances the oriented normal estimation results of both the end-to-end methods~\cite{li2023shs} and the neural gradient learning methods~\cite{li2023neural,li2024neuralgf} by substituting the initialized normals of PCA and MST with the ones estimated by other methods. More refinement details and analysis are reported in the \emph{Supplementary Material}.

\begin{table}[t]
\centering
\small
\setlength{\tabcolsep}{1pt}
\renewcommand\arraystretch{1}
\caption{Statistical results of unoriented CND on the real-world SceneNN dataset.}
\begin{tabular}{c|cccccc}
\hline
Method  & Ours          & \multicolumn{1}{l}{SHS-Net} & \multicolumn{1}{l}{Du~\etal}  & \multicolumn{1}{l}{HSurf-Net} & \multicolumn{1}{l}{GraphFit} & \multicolumn{1}{l}{DeepFit} \\ \hline
Clean   & \textbf{6.08} & 6.14                        & 6.12                          & 6.09                          & 6.37                         & 8.58                        \\
Noise   & \textbf{9.76} & 10.32                       & 9.93                          & 10.66                         & 10.81                        & 11.51                       \\
Average & \textbf{7.92} & 8.23                        & 8.03                          & 8.38                          & 8.59                         & 10.04                       \\ \hline
\end{tabular}
\label{tab: scenennn}
\vskip -0.3cm
\end{table}

\begin{table*}[htbp]
\centering
\small
\setlength{\tabcolsep}{3.5pt}
\renewcommand\arraystretch{1}
\caption{Quantitative comparisons of oriented normal estimation in terms of RMSE and CND on the PCPNet dataset.}
\begin{tabular}{l|cccc|cc|c|cccc|cc|c}
\hline
{\multirow{3}{*}{Method}}&\multicolumn{7}{c|}{{{RMSE}}}& \multicolumn{7}{c}{{CND}}  \\ \cline{2-15}
& \multicolumn{4}{c|}{{Noise ($\sigma$)}}& \multicolumn{2}{c|}{{Density}}& \multirow{2}{*}{{Ave.}}       & \multicolumn{4}{c|}{{Noise ($\sigma$)}}& \multicolumn{2}{c|}{{Density}} & \multirow{2}{*}{{Ave.}}  \\
&None          & 0.12\%        & 0.6\%          & \multicolumn{1}{c|}{1.2\%}          & Stripe       & \multicolumn{1}{c|}{Grad.}      & & None          & 0.12\%        & 0.6\%          & \multicolumn{1}{c|}{1.2\%}          & Stripe    & \multicolumn{1}{c|}{Grad.}      & \\ \hline
\hline
PCA~\cite{hoppe1992surface}+MST~\cite{hoppe1992surface} & 26.97         & 26.40          & 44.24          & \multicolumn{1}{c|}{44.55}                         & 26.84          & \multicolumn{1}{c|}{31.32}                         & 33.40                     & 26.97         & 26.43          & 43.03          & \multicolumn{1}{c|}{42.30}                         & 26.84          & \multicolumn{1}{c|}{31.32}                         & 32.82                     \\
PCA~\cite{hoppe1992surface}+SNO~\cite{schertler2017towards} & 18.87         & 23.79          & 31.83          & \multicolumn{1}{c|}{40.28}                         & 24.12          & \multicolumn{1}{c|}{26.71}                         & 27.60                     & 18.87         & 23.75          & 29.01          & \multicolumn{1}{c|}{35.11}                         & 24.12          & \multicolumn{1}{c|}{26.71}                         & 26.26                     \\
PCA~\cite{hoppe1992surface}+ODP~\cite{metzer2021orienting} & 27.10         & 26.02          & 34.97          & \multicolumn{1}{c|}{51.78}                         & 27.32          & \multicolumn{1}{c|}{21.38}                         & 31.43                     & 27.10         & 26.00          & 32.10          & \multicolumn{1}{c|}{47.55}                         & 27.32          & \multicolumn{1}{c|}{21.38}                         & 30.24                     \\
AdaFit~\cite{zhu2021adafit}+MST~\cite{hoppe1992surface} & 33.79         & 35.51          & 46.24          & \multicolumn{1}{c|}{52.88}                         & 32.53          & \multicolumn{1}{c|}{41.80}                         & 40.46                     & 33.79         & 35.46          & 44.68          & \multicolumn{1}{c|}{50.42}                         & 32.53          & \multicolumn{1}{c|}{41.80}                         & 39.78                     \\
AdaFit~\cite{zhu2021adafit}+SNO~\cite{schertler2017towards} & 27.81         & 27.13          & 40.56          & \multicolumn{1}{c|}{48.79}                         & 26.34          & \multicolumn{1}{c|}{31.77}                         & 33.73                     & 27.81         & 26.98          & 38.17          & \multicolumn{1}{c|}{43.21}                         & 26.34          & \multicolumn{1}{c|}{31.77}                         & 32.38                     \\
AdaFit~\cite{zhu2021adafit}+ODP~\cite{metzer2021orienting} & 27.66         & 27.52          & 35.66          & \multicolumn{1}{c|}{51.43}                         & 24.72          & \multicolumn{1}{c|}{20.46}                         & 31.24                     & 27.66         & 27.37          & 32.55          & \multicolumn{1}{c|}{44.06}                         & 24.72          & \multicolumn{1}{c|}{20.46}                         & 29.47                     \\
HSurf-Net~\cite{li2022hsurf}+MST~\cite{hoppe1992surface} & 35.87         & 32.86          & 55.21          & \multicolumn{1}{c|}{55.52}                         & 44.16          & \multicolumn{1}{c|}{32.60}                         & 42.71                     & 35.87         & 32.80          & 53.94          & \multicolumn{1}{c|}{53.39}                         & 44.16          & \multicolumn{1}{c|}{32.60}                         & 42.13                     \\
HSurf-Net~\cite{li2022hsurf}+SNO~\cite{schertler2017towards} & 31.53         & 33.24          & 44.71          & \multicolumn{1}{c|}{51.28}                         & 32.44          & \multicolumn{1}{c|}{39.23}                         & 38.74                     & 31.53         & 33.02          & 41.01          & \multicolumn{1}{c|}{45.31}                         & 32.44          & \multicolumn{1}{c|}{39.23}                         & 37.09                     \\
HSurf-Net~\cite{li2022hsurf}+ODP~\cite{metzer2021orienting} & 27.39         & 25.21          & 36.07          & \multicolumn{1}{c|}{49.64}                         & 25.18          & \multicolumn{1}{c|}{20.07}                         & 30.59                     & 27.39         & 25.01          & 32.51          & \multicolumn{1}{c|}{41.85}                         & 25.18          & \multicolumn{1}{c|}{20.07}                         & 28.67                     \\
PCPNet~\cite{guerrero2018pcpnet} & 70.44         & 69.82          & 67.02          & \multicolumn{1}{c|}{67.19}                         & 63.07          & \multicolumn{1}{c|}{56.79}                         & 65.72                     & 70.44         & 69.75          & 65.79          & \multicolumn{1}{c|}{64.02}                         & 63.07          & \multicolumn{1}{c|}{56.79}                         & 64.98                     \\
SHS-Net~\cite{li2023shs} & 10.23         & 14.26          & 26.68          & \multicolumn{1}{c|}{37.28}                         & 14.74          & \multicolumn{1}{c|}{16.79}                         & 20.00                     & 10.23         & 13.95          & 21.06          & \multicolumn{1}{c|}{29.01}                         & 14.74          & \multicolumn{1}{c|}{16.79}                         & 17.63                     \\
NGLO~\cite{li2023neural} & 13.52         & 18.77          & 35.72          & \multicolumn{1}{c|}{38.23}                         & 24.86          & \multicolumn{1}{c|}{\textbf{9.03}}                 & 23.36                     & 13.52         & 18.43          & 29.83          & \multicolumn{1}{c|}{28.64}                         & 24.86          & \multicolumn{1}{c|}{\textbf{9.03}}                 & 20.72                     \\
NeuralGF~\cite{li2024neuralgf} & 10.74         & 14.61          & 26.78          & \multicolumn{1}{c|}{\textbf{33.90}}                & \textbf{12.08} & \multicolumn{1}{c|}{12.88}                         & 18.50                     & 10.74         & 14.25          & 19.81          & \multicolumn{1}{c|}{\textbf{24.51}}                & \textbf{12.08} & \multicolumn{1}{c|}{12.88}                         & 15.71                     \\
\cellcolor{boxbody}{Ours}                                            & \cellcolor{boxbody}{\textbf{8.98}} & \cellcolor{boxbody}{\textbf{11.19}} & \cellcolor{boxbody}{\textbf{26.32}} & \cellcolor{boxbody}{36.86} & \cellcolor{boxbody}{12.93}          & \cellcolor{boxbody}{13.14} & \cellcolor{boxbody}{\textbf{18.24}}            & \cellcolor{boxbody}{\textbf{8.98}} & \cellcolor{boxbody}{\textbf{10.83}} & \cellcolor{boxbody}{\textbf{19.14}} & \cellcolor{boxbody}{27.27} & \cellcolor{boxbody}{12.93}          & \cellcolor{boxbody}{13.14} & \cellcolor{boxbody}{\textbf{15.38}}            \\ \hline
\end{tabular}
\label{tab: pcpnet_o}
\end{table*}

\begin{table*}[htbp]
\centering
\small
\setlength{\tabcolsep}{3.5pt}
\renewcommand\arraystretch{1}
\caption{Quantitative comparisons of oriented normal estimation regarding RMSE and CND on the FamousShape dataset.}
\begin{tabular}{l|cccc|cc|c|cccc|cc|c}
\hline
{\multirow{3}{*}{Method}}&\multicolumn{7}{c|}{{{RMSE}}}& \multicolumn{7}{c}{{CND}}  \\ \cline{2-15}
& \multicolumn{4}{c|}{{Noise ($\sigma$)}}& \multicolumn{2}{c|}{{Density}}& \multirow{2}{*}{{Ave.}}       & \multicolumn{4}{c|}{{Noise ($\sigma$)}}& \multicolumn{2}{c|}{{Density}} & \multirow{2}{*}{{Ave.}}  \\
&None          & 0.12\%        & 0.6\%          & \multicolumn{1}{c|}{1.2\%}          & Stripe       & \multicolumn{1}{c|}{Grad.}      & & None          & 0.12\%        & 0.6\%          & \multicolumn{1}{c|}{1.2\%}          & Stripe    & \multicolumn{1}{c|}{Grad.}      & \\ \hline
\hline
PCA~\cite{hoppe1992surface}+MST~\cite{hoppe1992surface} & 37.78          & 45.05          & 55.35          & \multicolumn{1}{c|}{75.47}                                  & 43.57          & \multicolumn{1}{c|}{47.87}                         & 50.85                     & 37.78          & 44.95          & 52.82          & \multicolumn{1}{c|}{72.09}                                  & 43.57          & \multicolumn{1}{c|}{47.87}                         & 49.85                     \\
PCA~\cite{hoppe1992surface}+SNO~\cite{schertler2017towards} & 36.24          & 37.31          & 43.59          & \multicolumn{1}{c|}{72.82}                                  & 44.39          & \multicolumn{1}{c|}{43.15}                         & 46.25                     & 36.24          & 37.17          & 41.32          & \multicolumn{1}{c|}{69.73}                                  & 44.39          & \multicolumn{1}{c|}{43.15}                         & 45.33                     \\
PCA~\cite{hoppe1992surface}+ODP~\cite{metzer2021orienting} & 35.91          & 29.34          & 44.51          & \multicolumn{1}{c|}{79.34}                                  & 51.66          & \multicolumn{1}{c|}{35.68}                         & 46.07                     & 35.91          & 29.25          & 41.29          & \multicolumn{1}{c|}{76.98}                                  & 51.66          & \multicolumn{1}{c|}{35.68}                         & 45.13                     \\
AdaFit~\cite{zhu2021adafit}+MST~\cite{hoppe1992surface} & 33.36          & 49.70          & 95.52          & \multicolumn{1}{c|}{65.04}                                  & 70.17          & \multicolumn{1}{c|}{55.31}                         & 61.52                     & 33.36          & 49.40          & 95.33          & \multicolumn{1}{c|}{60.96}                                  & 70.17          & \multicolumn{1}{c|}{55.31}                         & 60.75                     \\
AdaFit~\cite{zhu2021adafit}+SNO~\cite{schertler2017towards} & 34.82          & 40.02          & 61.37          & \multicolumn{1}{c|}{68.13}                                  & 38.71          & \multicolumn{1}{c|}{40.61}                         & 47.28                     & 34.82          & 39.77          & 56.92          & \multicolumn{1}{c|}{64.08}                                  & 38.71          & \multicolumn{1}{c|}{40.61}                         & 45.82                     \\
AdaFit~\cite{zhu2021adafit}+ODP~\cite{metzer2021orienting} & 36.38          & 32.77          & 42.49          & \multicolumn{1}{c|}{79.30}                                  & 38.10          & \multicolumn{1}{c|}{46.55}                         & 45.93                     & 36.38          & 32.51          & 38.24          & \multicolumn{1}{c|}{76.67}                                  & 38.10          & \multicolumn{1}{c|}{46.55}                         & 44.74                     \\
HSurf-Net~\cite{li2022hsurf}+MST~\cite{hoppe1992surface} & 57.83          & 43.78          & 79.42          & \multicolumn{1}{c|}{75.44}                                  & 70.97          & \multicolumn{1}{c|}{39.64}                         & 61.18                     & 57.83          & 43.50          & 78.58          & \multicolumn{1}{c|}{72.84}                                  & 70.97          & \multicolumn{1}{c|}{39.64}                         & 60.56                     \\
HSurf-Net~\cite{li2022hsurf}+SNO~\cite{schertler2017towards} & 40.38          & 38.91          & 64.51          & \multicolumn{1}{c|}{66.21}                                  & 49.70          & \multicolumn{1}{c|}{44.17}                         & 50.65                     & 40.38          & 38.56          & 61.21          & \multicolumn{1}{c|}{62.48}                                  & 49.70          & \multicolumn{1}{c|}{44.17}                         & 49.42                     \\
HSurf-Net~\cite{li2022hsurf}+ODP~\cite{metzer2021orienting} & 38.87          & 36.12          & 47.65          & \multicolumn{1}{c|}{74.88}                                  & 48.53          & \multicolumn{1}{c|}{46.17}                         & 48.70                     & 38.87          & 35.90          & 43.96          & \multicolumn{1}{c|}{71.00}                                  & 48.53          & \multicolumn{1}{c|}{46.17}                         & 47.40                     \\
PCPNet~\cite{guerrero2018pcpnet}              & 69.88          & 68.59          & 68.87          & \multicolumn{1}{c|}{68.47}                                  & 69.34          & \multicolumn{1}{c|}{73.91}                         & 69.84                     & 69.88          & 68.57          & 67.56          & \multicolumn{1}{c|}{63.33}                                  & 69.34          & \multicolumn{1}{c|}{73.91}                         & 68.76                     \\
SHS-Net~\cite{li2023shs}                  & 20.69          & 27.10          & 42.49          & \multicolumn{1}{c|}{53.66}                                  & 23.76          & \multicolumn{1}{c|}{25.19}                         & 32.14                     & 20.69          & 26.70          & 37.50          & \multicolumn{1}{c|}{43.95}                                  & 23.76          & \multicolumn{1}{c|}{25.19}                         & 29.63                     \\
NGLO~\cite{li2023neural}              & \textbf{13.26} & \textbf{19.97} & 38.63          & \multicolumn{1}{c|}{52.67}                                  & 26.06          & \multicolumn{1}{c|}{\textbf{13.49}}                & 27.35                     & \textbf{13.26} & \textbf{19.39} & 33.18          & \multicolumn{1}{c|}{43.77}                                  & 26.06          & \multicolumn{1}{c|}{\textbf{13.49}}                & 24.86                     \\
NeuralGF~\cite{li2024neuralgf}            & 16.85          & 20.38          & \textbf{36.98} & \multicolumn{1}{c|}{51.31}                                  & 17.33          & \multicolumn{1}{c|}{17.16}                         & 26.67                     & 16.85          & 19.81          & 32.25          & \multicolumn{1}{c|}{41.59}                                  & 17.33          & \multicolumn{1}{c|}{17.16}                         & 24.17                     \\
\cellcolor{boxbody}{Ours}                                            & \cellcolor{boxbody}{15.97} & \cellcolor{boxbody}{20.68} & \cellcolor{boxbody}{38.52} & \cellcolor{boxbody}{\textbf{50.89}} & \cellcolor{boxbody}{\textbf{17.23}} & \cellcolor{boxbody}{15.65} & \cellcolor{boxbody}{\textbf{26.49}}          & \cellcolor{boxbody}{15.97} & \cellcolor{boxbody}{20.09} & \cellcolor{boxbody}{\textbf{32.10}} & \cellcolor{boxbody}{\textbf{39.52}} & \cellcolor{boxbody}{\textbf{17.23}} & \cellcolor{boxbody}{15.65} & \cellcolor{boxbody}{\textbf{23.43}}             \\ \hline
\end{tabular}
\label{tab: famous_o}
\vskip -0.3cm
\end{table*}



\subsubsection{Computational Cost Comparison}

To further demonstrate the efficiency superiority of our methods, we conduct a computational time evaluation against the neural gradient learning methods~\cite{li2023neural,li2024neuralgf} on the FamousShape dataset. In this dataset, each test point cloud has 100K sampled points, and all experiments are executed on a single NVIDIA A100 GPU. The statistic time of our method contains both the initialization part of PCA and MST and the network inference part. As discussed in Sec.~\ref{rw} and Sec.~\ref{re}, the neural gradient learning methods necessitate re-training from scratch to adapt to the geometry of a new test point cloud, and thus we consider both the training and inference time of the neural gradient learning methods for a fair comparison. The statistical results in Table~\ref{tab: time} suggest that our method substantially reduces the normal estimation time in contrast to the neural gradient learning methods. Moreover, the proposed refinement framework enhances estimation results with minimal computational overhead. We report additional results and experimental configurations in the \emph{Supplementary Material}.

\begin{table}[!t]
\centering
\small
\setlength{\tabcolsep}{2.4pt}
\renewcommand\arraystretch{1}
\caption{Statistical results of oriented normal estimation by other methods using our refinement strategy.}
\begin{tabular}{l|cccc|cc|c}
\hline
\multicolumn{1}{c|}{\multirow{2}{*}{Method}} & \multicolumn{4}{c|}{Noise ($\sigma$)}                                     & \multicolumn{2}{c|}{Density}    & \multirow{2}{*}{Ave.} \\
\multicolumn{1}{c|}{}                          & None          & 0.12\%         & 0.6\%          & 1.2\%          & Stripe         & Grad.       &                          \\ \hline
SHS-Net                                        & 10.23         & 13.95          & 21.06          & 29.01          & 14.74          & 16.79          & 17.63                    \\
SHS-Net+Ours                                   & \textbf{9.22} & \textbf{13.36} & \textbf{19.04} & \textbf{27.65} & \textbf{13.71} & \textbf{15.84} & \textbf{16.47}           \\ \hline
NGLO                                           & 13.52         & 18.43          & 29.83          & 28.64          & 24.86          & \textbf{9.03}  & 20.72                    \\
NGLO+Ours                                      & \textbf{8.76} & \textbf{15.20} & \textbf{21.96} & \textbf{25.72} & \textbf{14.32} & 11.10          & 16.18                    \\ \hline
NGF                                            & 10.74         & 14.25          & 19.81          & \textbf{24.51} & 12.08          & 12.88          & 15.71                    \\
NGF+Ours                                       & \textbf{7.75} & \textbf{11.14} & \textbf{17.14} & 25.59          & \textbf{10.24} & \textbf{11.47} & \textbf{13.89}           \\ \hline
\end{tabular}
\label{tab: refine}
\end{table}

\subsection{Ablation Studies}

\subsubsection{The architecture of OCMG-Net}
The OCMG-Net architecture consists of four important components: \emph{Orientation Framework}, \emph{Multi-scale Local Feature Aggregation}, \emph{Hierarchical Geometric Information Fusion}, and \emph{Unoriented Normal Estimation Module}. We delve into the functionalities of these components on the PCPNet dataset. Additionally, we perform ablation experiments on the loss functions and scale parameters to further understand their impact and effectiveness within the network.
\begin{table}[!t]
\centering
\small
\setlength{\tabcolsep}{1.5pt}
\renewcommand\arraystretch{1}
\caption{Efficiency comparisons (seconds per 100K
points).}
\begin{tabular}{c|ccccc}
\hline
\multirow{2}{*}{Method} & Ours          & Ours       & Ours                   & \multirow{2}{*}{NGLO}   & \multirow{2}{*}{Neural-GF} \\
                        & MST Init. \& DA & MST Init. & Base                   &                         &                            \\ \hline
\multirow{2}{*}{Time (s)}   & \textbf{64.37}         & \textbf{64.29}      & \multirow{2}{*}{\textbf{59.83}} & \multirow{2}{*}{445.86} & \multirow{2}{*}{350.68}    \\
                        & +4.46+0.08    & +4.46      &                        &                         &                            \\ \hline
\end{tabular}
\label{tab: time}
\vskip -0.3cm
\end{table}\\
\textbf{(a).} To validate the effectiveness of the proposed refinement framework, experiments are conducted without MST initialization and data augmentation in the normal feature space. As shown in Table~\ref{tab: ablation}(a), the orientation initialization and data augmentation can significantly enhance the network's sign orientation capabilities. Furthermore, we analyze the importance of global patch feature, global point cloud feature, and the normal feature utilized in the orientation sign refinement branch. The results in Table~\ref{tab: ablation}(a) demonstrate the critical nature of these features for oriented normal estimation.\\
\textbf{(b).} In the Multi-scale Local Feature Aggregation, we capture the local structure using two scales and integrate them by Attention Feature Fusion. Table~\ref{tab: ablation}(b) reports the results of 1) without Local Feature Extraction; 2) with Single-scale Local Feature Extraction only, and 3) direct integration of multi-scale local features using an MLP instead of an attention module. As observed, the incorporation of multi-scale local features with Attention Feature Fusion significantly enhances the depiction of input patches and point clouds, resulting in more accurate unoriented and oriented normal estimation results compared to the other configurations.\\
\textbf{(c).} To validate the effectiveness of the Hierarchical Geometric Information Fusion, we carry out experiments using the model with a fixed global scale that is equivalent to the output scale of OCMG-Net. Additionally, we compare the results of the models without the global feature of the last scale or the local feature in the hierarchical architecture. Results shown in Table~\ref{tab: ablation}(c) demonstrate that hierarchical architecture is important for feature extraction. Further, the Hierarchical Geometric Information Fusion with cross-scale global feature and various local feature can also boost both the unoriented and oriented normal estimation performance.\\
\textbf{(d).} Table~\ref{tab: ablation}(d) shows the ablation results of the unoriented normal estimation part, suggesting the effectiveness of Position Feature Fusion and Weighted Normal Prediction.\\
\textbf{(e).} Additionally, we perform ablation studies on the loss functions. The results reported in Table~\ref{tab: ablation}(e) indicate that both the $\boldsymbol{z}$-direction loss for QSTN and the contrastive loss for data augmentation contribute to enhancing both unoriented and oriented normal estimation results.
Besides, the CND-modified loss can guide the trained network towards obtaining more accurate normals, particularly in noisy regions.\\
\textbf{(f).} The results of different scale parameter settings are reported in Table~\ref{tab: ablation}(f). It is observed that thanks to the scale adaptability of the proposed feature extraction architecture, alterations in the scales of the input patch $N_P$ and point cloud $N_D$ have minimal impact on the normal estimation results compared to when the proposed modules are not utilized. Moreover, our chosen downsampling factors $\rho_{\boldsymbol{P}}=\{2/3, 2/3, 2/3, 1\}$ are demonstrated to outperform other selections, further validating the effectiveness of our approach in handling scale variations.



\begin{table*}[t]
\centering
\small
\setlength{\tabcolsep}{6.5pt}
\renewcommand\arraystretch{1}
\caption{Ablation studies with the (a) Orientation Framework; (b) Multi-scale Local Feature Aggregation; (c) Hierarchical Architecture; (d) Unoriented Normal Estimation Module; (e) Loss Function; and (f) Scale Parameters.}
\begin{tabular}{cc|cccc|cc|c|c}
\hline
\multicolumn{2}{c|}{\multirow{2}{*}{Ablation Studies}}            & \multicolumn{4}{c|}{Noise ($\sigma$)}                                     & \multicolumn{2}{c|}{Density}    & Oirented       & Unoriented    \\
\multicolumn{2}{c|}{}                                     & None          & 0.12\%         & 0.60\%         & 1.20\%         & Stripe         & Grad.          & Average        & Average       \\ \hline
\multirow{5}{*}{(a)} & w/o MST Initialization             & 10.69         & 12.28          & 19.18          & 27.65          & 15.67          & 15.12          & 16.77          & 8.17          \\
                     & w/o Data Augmentation             & 9.28          & 11.68          & 19.21          & 27.41          & 13.51          & 13.93          & 15.83          & 8.15          \\
                     & w/o Patch Feature                  & 13.10         & 15.15          & 21.32          & 27.42          & 15.65          & 16.29          & 18.15          & 8.19          \\
                     & w/o Point Cloud Feature            & 58.99         & 42.84          & 45.61          & 36.05          & 59.31          & 69.89          & 52.11          & 8.36          \\
                     & w/o Normal Feature                 & 14.03         & 16.90          & 22.64          & 29.31          & 17.68          & 17.36          & 19.65          & 8.18          \\ \hline
\multirow{3}{*}{(b)} & w/o Local Feature Extration        & 13.08         & 16.07          & 21.00          & 31.48          & 18.82          & 13.98          & 19.07          & 8.83          \\
                     & w/ Single Local Feature            & 11.87         & 13.17          & 20.09          & 27.66          & 13.97          & 15.09          & 16.98          & 8.45          \\
                     & w/o Attentional Feature Fusion     & 10.01         & 11.25          & 20.02          & 27.55          & 13.14          & 13.62          & 15.93          & 8.31          \\ \hline
\multirow{3}{*}{(c)} & w/o Hierarchical Architecture      & 14.38         & 17.79          & 22.58          & 31.69          & 20.92          & 20.61          & 21.33          & 9.87          \\
                     & w/o Cross-scale Global Feature     & 10.45         & 14.59          & 19.69          & 28.93          & 15.19          & 17.17          & 17.67          & 8.31          \\
                     & w/o Various Local Feature          & 9.60          & 11.83          & 20.78          & 28.77          & 14.12          & 15.88          & 16.83          & 8.72          \\ \hline
\multirow{2}{*}{(d)} & w/o Position Feature Fusion        & 10.11         & 11.31          & 19.94          & 27.97          & 12.96          & 14.61          & 16.15          & 8.46          \\
                     & w/o Weighted Normal Prediction     & 9.38          & 11.24          & 19.37          & 28.07          & 13.11          & 14.31          & 15.91          & 8.63          \\ \hline
\multirow{3}{*}{(e)} & w/o  $\mathbf{z}$-direction Loss              & 9.13          & 11.18          & 20.62          & 26.87          & 14.23          & 14.69          & 16.12          & 8.51          \\
                     & w/o  Contrastive Loss              & 9.41          & 11.05          & 19.30          & 27.44          & 13.07          & 13.37          & 15.61          & 8.14          \\
                     & w/o  CND-Modified Loss             & 9.21          & 11.38          & 20.41          & 28.47          & 12.98          & 14.68          & 16.19          & 8.48          \\ \hline
\multirow{6}{*}{(f)} & $N_P$ = 600                        & 9.01          & 11.07          & 19.58          & 28.79          & \textbf{12.93} & \textbf{12.73} & 15.69          & 8.23          \\
                     & $N_P$ = 800                        & 9.12          & \textbf{10.80} & \textbf{19.10} & \textbf{26.80} & 13.43          & 13.42          & \textbf{15.45} & 8.14          \\
                     & $N_D$ = 1000                       & 9.02          & 10.90          & 19.18          & 27.92          & 13.39          & 13.60          & 15.67          & \textbf{8.12} \\
                     & $N_D$ = 1400                       & 9.02          & 10.95          & 19.39          & 27.37          & 12.96          & \textbf{12.98} & \textbf{15.45} & \textbf{8.12} \\
                     & $\rho_{\boldsymbol{P}}$ = \{1/3, 1/3, 1, 1\}      & \textbf{8.73} & 11.07          & 20.58          & 30.51          & 13.40          & 13.05          & 16.22          & 8.33          \\
                     & $\rho_{\boldsymbol{P}}$ = \{1/2, 1/2, 1/2, 1\}    & 9.41          & 12.58          & 23.21          & 30.97          & 14.77          & 13.70          & 17.44          & 8.71          \\ \hline
\cellcolor{boxbody}{\textbf{}}&\cellcolor{boxbody}{{Ours}}                      & \cellcolor{boxbody}{\textbf{8.98}} & \cellcolor{boxbody}{\textbf{10.83}} & \cellcolor{boxbody}{\textbf{19.14}} & \cellcolor{boxbody}{\textbf{27.27}} & \cellcolor{boxbody}{\textbf{12.93}} & \cellcolor{boxbody}{13.14}          & \cellcolor{boxbody}{\textbf{15.38}} & \cellcolor{boxbody}{\textbf{8.12}} \\ \hline
\end{tabular}
\label{tab: ablation}
\end{table*}

\begin{table*}[t]
\centering
\small
\setlength{\tabcolsep}{8pt}
\renewcommand\arraystretch{1}
\caption{Network training with (\checkmark) or without (\ding{55}) the CND-modified loss function in both oriented and unoriented settings using the PCPNet dataset.}
\begin{tabular}{c|cc|cc|cc|cc|cc|cc}
\hline
\multirow{2}{*}{Method} & \multicolumn{4}{c|}{Oriented}                                                 & \multicolumn{8}{c}{Unoriented}                                                                                                                                   \\ \cline{2-13} 
                        & \multicolumn{2}{c|}{Ours}                   & \multicolumn{2}{c|}{SHS-Net}    & \multicolumn{2}{c|}{Ours}                           & \multicolumn{2}{c|}{SHS-Net}                & \multicolumn{2}{c|}{HSurf-Net} & \multicolumn{2}{c}{DeepFit} \\ \hline
$\mathcal{L}_{\mathrm{CND}}$                     & \checkmark             & \ding{55}      & \checkmark              &                & \checkmark              & \ding{55}             & \checkmark              & \ding{55}     & \checkmark & \ding{55}      & \checkmark          &   \ding{55}              \\ \hline
No Noise                & \textbf{8.96}  & \multicolumn{1}{c|}{9.21}  & 10.92          & \textbf{10.23} & \textbf{3.48}  & \multicolumn{1}{c|}{3.50}          & \textbf{3.75}  & \multicolumn{1}{c|}{3.82}  & 3.98 & \multicolumn{1}{c|}{3.91}  &  6.63      & 6.55           \\
Low Noise               & \textbf{10.83} & \multicolumn{1}{c|}{11.38} & \textbf{13.42} & 13.95          & \textbf{8.30}  & \multicolumn{1}{c|}{8.47}          & \textbf{8.35}  & \multicolumn{1}{c|}{8.43}  & \textbf{8.47} & \multicolumn{1}{c|}{8.54}  & \textbf{9.05} & 9.24           \\
Med Noise               & \textbf{19.14} & \multicolumn{1}{c|}{20.41} & \textbf{19.63} & 21.06          & \textbf{12.40} & \multicolumn{1}{c|}{13.17}         & \textbf{12.76} & \multicolumn{1}{c|}{13.49} & \textbf{12.86} & \multicolumn{1}{c|}{13.23} & \textbf{13.86} & 14.09          \\
High Noise              & \textbf{27.27} & \multicolumn{1}{c|}{28.57} & \textbf{26.58} & 29.01          & \textbf{16.10} & \multicolumn{1}{c|}{17.18}         & \textbf{16.33} & \multicolumn{1}{c|}{17.25} & \textbf{16.65} & \multicolumn{1}{c|}{16.85} & \textbf{19.03} & 19.35          \\
Stripes                 & \textbf{12.91} & \multicolumn{1}{c|}{12.98} & 15.13          & \textbf{14.74} & \textbf{4.16}  & \multicolumn{1}{c|}{4.32}          & \textbf{4.45}  & \multicolumn{1}{c|}{4.53}  & 4.59 & \multicolumn{1}{c|}{4.56}  & 7.77       & 7.72           \\
Gradients               & \textbf{13.14} & \multicolumn{1}{c|}{14.68} & \textbf{16.58} & 16.79          & 4.25           & \multicolumn{1}{c|}{\textbf{4.24}} & \textbf{4.41}  & \multicolumn{1}{c|}{4.49}  & \textbf{4.52} & \multicolumn{1}{c|}{4.55}  & 7.31       & 7.29           \\ \hline
Average                 & \textbf{15.38} & \multicolumn{1}{c|}{16.19} & \textbf{17.04} & 17.63          & \textbf{8.12}  & \multicolumn{1}{c|}{8.48}          & \textbf{8.34}  & \multicolumn{1}{c|}{8.67}  & \textbf{8.51} & \multicolumn{1}{c|}{8.61}  & \textbf{10.61} & 10.71          \\ \hline
\end{tabular}
\label{tab: cnd}
\vskip -0.3cm
\end{table*}

\subsubsection{Generalization of the CND-modified Loss Function}
Next, we conduct experiments on the PCPNet dataset to demonstrate the effectiveness and generalization of the newly introduced CND-modified loss function by comparing the results with and without its integration. We evaluate representative methods for both unoriented and oriented normal estimation, including the end-to-end oriented normal estimation method SHS-Net~\cite{li2023shs}, the deep surface fitting method DeepFit~\cite{ben2020deepfit}, as well as the regression method Hsurf-Net~\cite{li2022hsurf}, alongside our method. Table~\ref{tab: cnd} highlights the impact of the CND component, demonstrating its significant enhancement in normal estimation accuracy on the noisy point clouds across all method categories, encompassing both unoriented and oriented normal estimation.



\begin{figure*}[t]
\centering
\includegraphics[width=0.9\linewidth]{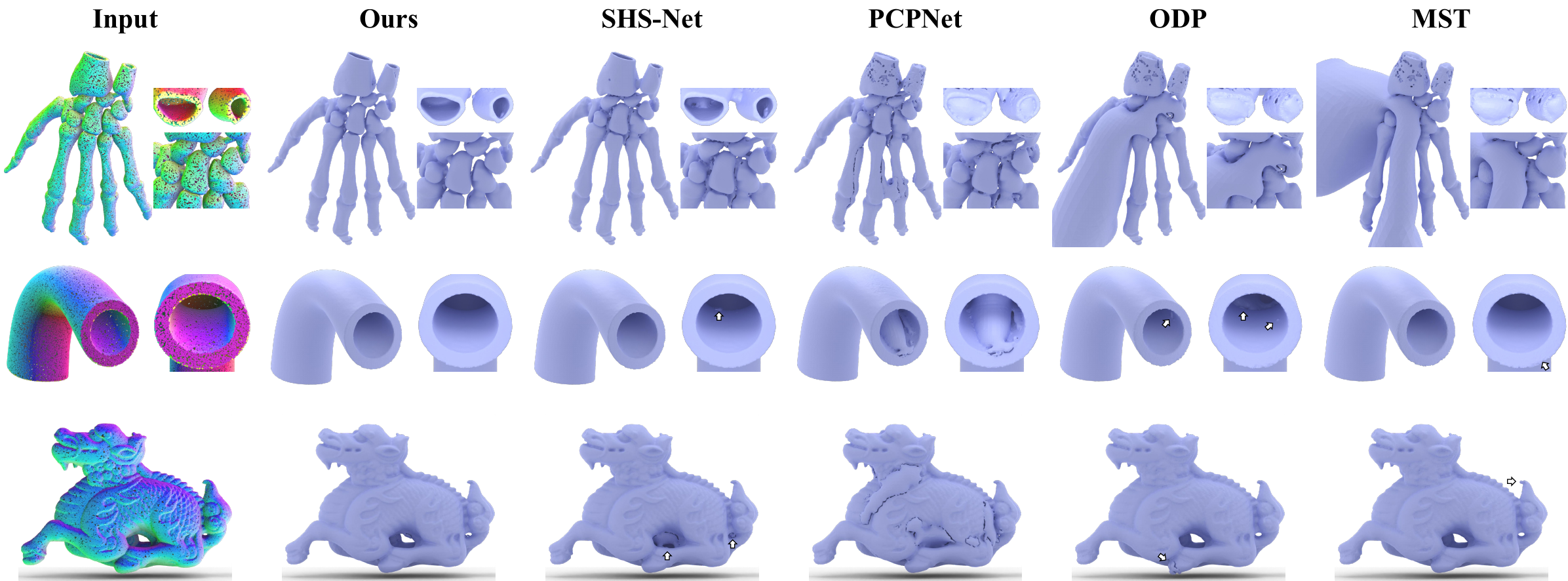}
\vskip -0.2cm
\caption{Qualitative comparisons of surface reconstruction from noise-free point clouds in the PCPNet and FamousShape datasets using oriented normals estimated by various methods.}
\label{fig: reconstruction}
\vskip -0.2cm
\end{figure*}

\begin{figure*}[!h]
\centering
\includegraphics[width=0.9\linewidth]{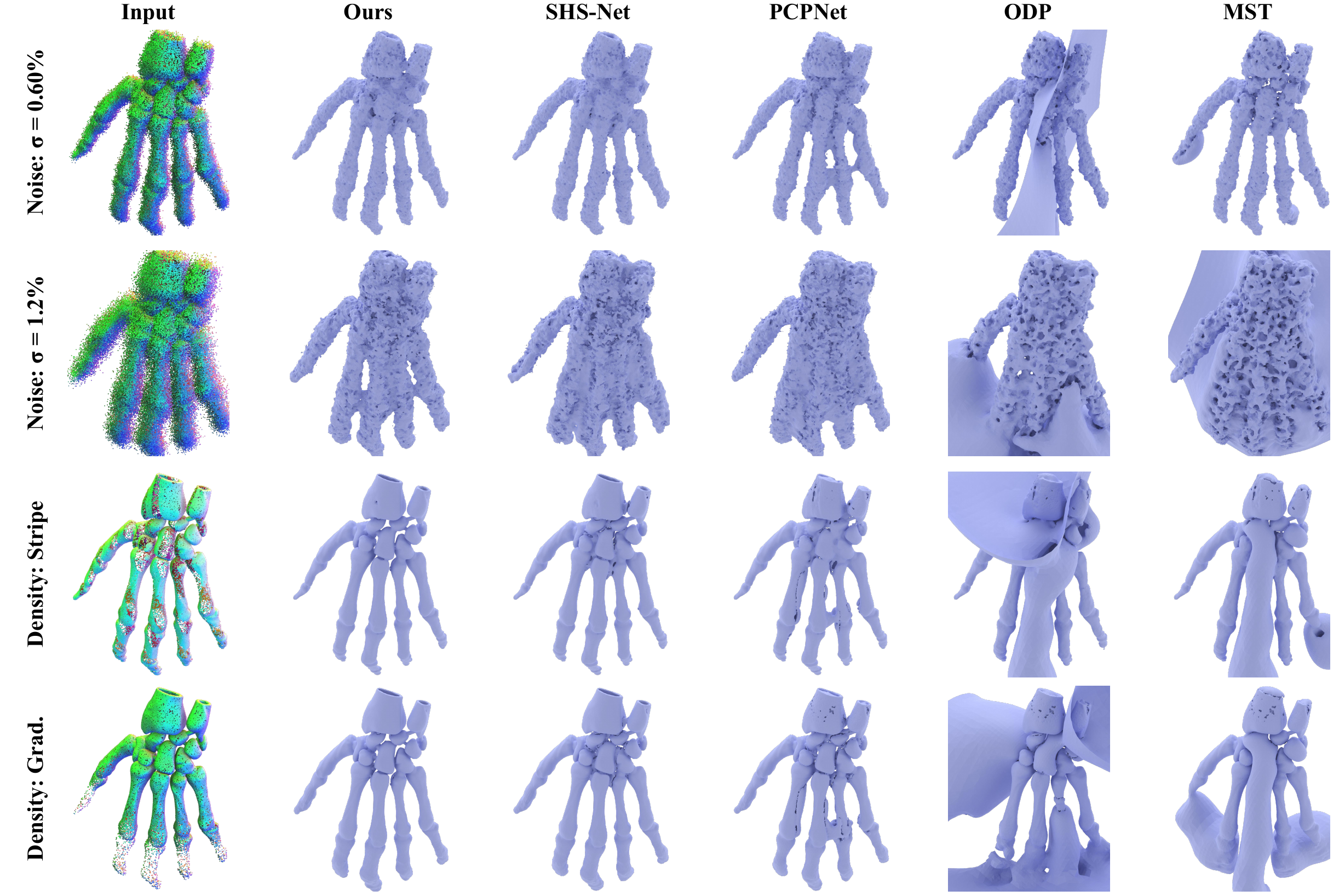}
\vskip -0.2cm
\caption{Qualitative comparisons of surface reconstruction from challenging point clouds under varying noise levels and density variations using oriented normals estimated by different approaches.}
\label{fig: reconstruction_noise}
\vskip -0.2cm
\end{figure*}

\subsubsection{Generalization of the Refinement Framework}
Moreover, we carry out experiments on the PCPNet dataset to verify the generalization of the proposed refinement framework. Using the representative end-to-end oriented normal estimation method SHS-Net~\cite{li2023shs} and our OCMG-Net as baselines, we compare the outcomes with and without MST initialization and data augmentation (DA) in the normal feature space. The results presented in Table~\ref{tab: refinr_app} demonstrate that both the initialization process and data augmentation significantly enhance the performance of end-to-end methods.


\begin{table}[t]
\centering
\small
\setlength{\tabcolsep}{6pt}
\renewcommand\arraystretch{1}
\caption{The end-to-end networks with or without our refinement framework on the PCPNet dataset.}
\begin{tabular}{c|ccc|ccc}
\hline
Method     & \multicolumn{3}{c|}{Ours}      & \multicolumn{3}{c}{SHS-Net}    \\ \hline
MST Init.  & \checkmark & \checkmark&  \ding{55}      & \checkmark & \checkmark&  \ding{55}      \\
DA         & \checkmark & \ding{55}       &   \ding{55}     & \checkmark & \ding{55}       &  \ding{55}      \\ \hline
No Noise   & \textbf{8.96}  & 9.28  & 10.69 & \textbf{9.04}  & 9.21  & 10.23 \\
Low Noise  & \textbf{10.83} & 11.68 & 12.28 & \textbf{12.75} & 13.07 & 13.95 \\
Med Noise  & \textbf{19.14} & 19.21 & 19.18 & \textbf{20.68} & 20.87 & 21.06 \\
High Noise & \textbf{27.27} & 27.41 & 27.65 & \textbf{28.71} & 28.76 & 29.01 \\
Stripes    & \textbf{12.91} & 13.51 & 15.67 & \textbf{13.28} & 14.01 & 14.74 \\
Gradients  & \textbf{13.14} & 13.93 & 15.12 & \textbf{14.03} & 14.73 & 16.79 \\ \hline
Average    & \textbf{15.38} & 15.83 & 16.77 & \textbf{16.42} & 16.78 & 17.63 \\ \hline
\end{tabular}
\label{tab: refinr_app}
\vskip -0.3cm
\end{table}

\subsection{Applications}
\subsubsection{Surface Reconstruction}
\nobf{Noise-Free Point Clouds.} 
We first showcase the application of our method on surface reconstruction using noise-free point clouds from the PCPNet and FamousShape datasets. In Fig.~\ref{fig: reconstruction}, we display the results of \emph{Poisson surface reconstruction}\cite{kazhdan2006poisson} utilizing the oriented normal vectors predicted by various competing approaches. We compare our method with the learning-based SHS-Net~\cite{li2023shs} and PCPNet~\cite{guerrero2018pcpnet}, as well as the traditional sign orientation methods ODP~\cite{metzer2021orienting} and MST~\cite{hoppe1992surface}, where unoriented normals are predicted using PCA~\cite{hoppe1992surface}. Our method excels in reconstructing areas with complex geometries and preserving thin structures such as holes and small gaps, outperforming other approaches. Moreover, the shapes reconstructed via our estimated normals exhibit minimal artifacts. We provide more reconstruction results in the \emph{Supplementary Material}.


\nobf{Noisy and Density-varied Point Clouds.} Next, we assess the robustness and generalization of our method to point clouds with varying noise levels and density fluctuations. The results in Fig.~\ref{fig: reconstruction_noise} validate that our method consistently delivers outstanding sign orientation results even as noise levels increasing, effectively mitigating the risk of thin gaps merging due to noise. Moreover, when confronted with point clouds with density variations, our method excels at reconstructing the original surface more accurately, especially in regions with sparsely sampled points. 

\begin{figure*}[t]
\centering
\includegraphics[width=0.9\linewidth]{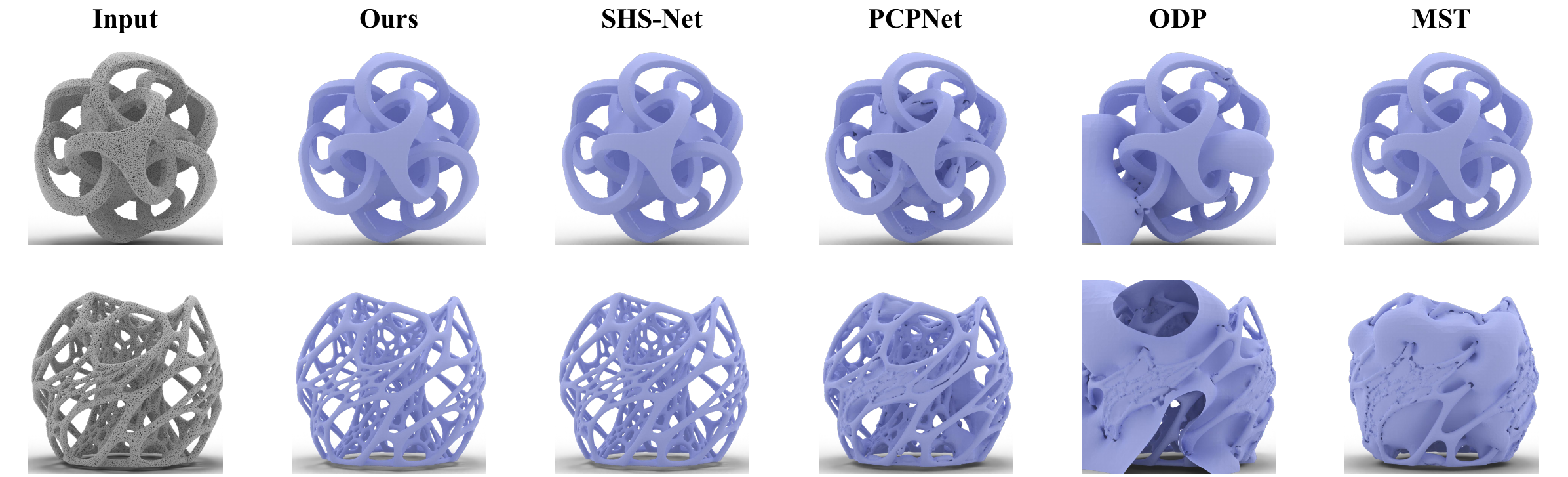}
\caption{Qualitative comparisons of surface reconstruction from challenging point clouds with complex typology using oriented normals estimated by different methods.}
\label{fig: reconstruction_c}
\vskip -0.05cm
\end{figure*}

\nobf{Point Clouds with Complex Topology.} Subsequently, we evaluate our method on point clouds~\cite{xu2023globally} characterized by \emph{nested structures} with highly complex topology, presenting significant challenges for normal orientation. As shown in Fig.~\ref{fig: reconstruction_c}, our method still achieves optimal surface reconstruction results without adhesion and oversmoothing, particularly in regions featuring adjacent surfaces or substantial curvature changes. 

\begin{figure*}[!h]
\centering
\includegraphics[width=0.9\linewidth]{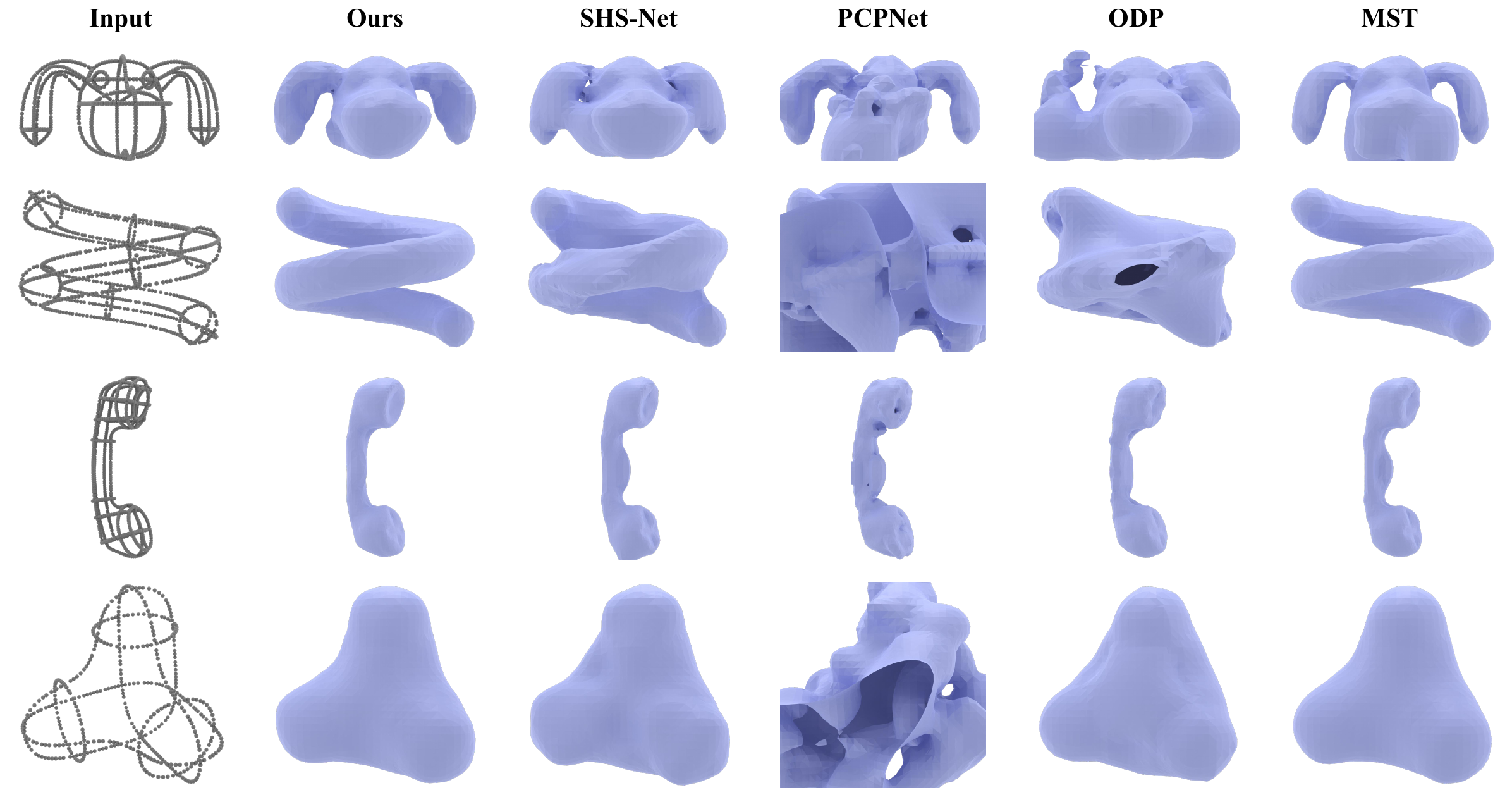}
\caption{Qualitative comparisons of surface reconstruction from wireframe point clouds with sparse and non-uniform sampling using oriented normals estimated by different methods.}
\label{fig: reconstruction_w}
\vskip -0.05cm
\end{figure*}

\begin{figure*}[!h]
\centering
\includegraphics[width=0.9\linewidth]{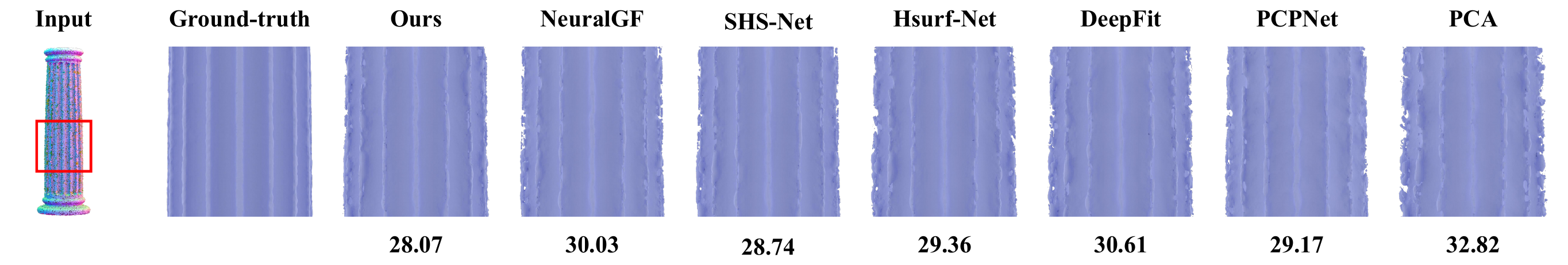}
\caption{Qualitative comparisons of point cloud denoising using normals estimated by different methods. The bottom presents the quantitative comparison results using the Chamfer Distance ($\times10^{-3}$).}
\label{fig: filter}
\vskip -0.05cm
\end{figure*}


\nobf{Wireframe Point Clouds.} 
To further highlight the efficacy of OCMG-Net on \emph{sparse data}, we utilize wireframe point clouds for comparison. Wireframe representations describe shapes through a compact skeletal structure with a quite limited number of points. Qualitative results in Fig.~\ref{fig: reconstruction_c} demonstrate that our method can avoid adhesion errors and reconstruct shapes more effectively from sparse and non-uniformly distributed data. Compared to learning-based methods~\cite{li2023shs, guerrero2018pcpnet}, OCMG-Net achieves superior sign orientation results and mitigates artifacts in reconstruction outcomes. In comparison to traditional methods~\cite{hoppe1992surface}, our method also delivers more accurate normal estimation and generates smoother reconstruction surfaces.


\subsubsection{Point Cloud Denoising} Point cloud denoising is another important downstream task following normal estimation, especially considering the inevitable presence of noise in real-world applications. We employ the filtering approach proposed in~\cite{lu2020low} to denoise point clouds using the estimated point normals obtained from various methods. As illustrated in Fig.~\ref{fig: filter}, in the local zoomed-in reconstruction view, the denoising results achieved using our estimated normals result in smoother surfaces and better preservation of sharp edges. Additionally, the reconstruction surfaces of the point cloud filtered using our estimated normals has minimal artifacts. We present more experimental details and qualitative comparisons in the \emph{Supplementary Material}.

\section{Limitations}
While our method has demonstrated remarkable accuracy in both unoriented and oriented normal estimation problems across diverse 3D models with improved efficiency, there are rooms where computational costs can be optimized further. Besides, the proposed OCMG-Net is a supervised learning method depending on annotated training data, as is the case with previous approaches. Therefore, in the future work, the architecture of the network can be further streamlined and it is highly desirable to delve into unsupervised frameworks.


\section{Conclusions}
In this work, we have presented an  in-depth analysis of oriented normal estimation and introduced the novel method OCMG-Net to achieve state-of-the-art performance across diverse datasets and scenarios.
Striving for a better balance between accuracy and efficiency, we propose a lightweight and soft refinement framework that acts as a \emph{universal enhancement} and fundamental module for common end-to-end approaches. To mitigate noise disturbances, we identify direction inconsistency in previous methods and introduce the CND metric to rectify this issue. This metric not only provides a more reasonable evaluation of existing methods but also enhances network training, significantly boosting network robustness against noise in both unoriented and oriented domains. 
Moreover, we present a novel architecture that enables the network to leverage geometric information more effectively, addressing scale selection ambiguity. This not only benefits normal estimation tasks but also contributes to the broader point cloud processing community.

Through extensive experiments, we validate the generalizability of our framework and the effectiveness of the newly defined CND loss function across various approaches. The experiments demonstrate that OCMG-Net outperforms competitors in both unoriented and oriented normal estimation settings in terms of both accuracy and robustness. Additionally, we showcase its versatility in real-world scenarios and downstream tasks, including point cloud denoising and surface reconstruction, where our method consistently delivers superior results.

\bibliographystyle{IEEEtran}
\bibliography{IEEEabrv, oriented_normal_estimation}
\vskip -1cm
\begin{IEEEbiography}
[{\includegraphics[width=1in,height=1.25in,clip,keepaspectratio]{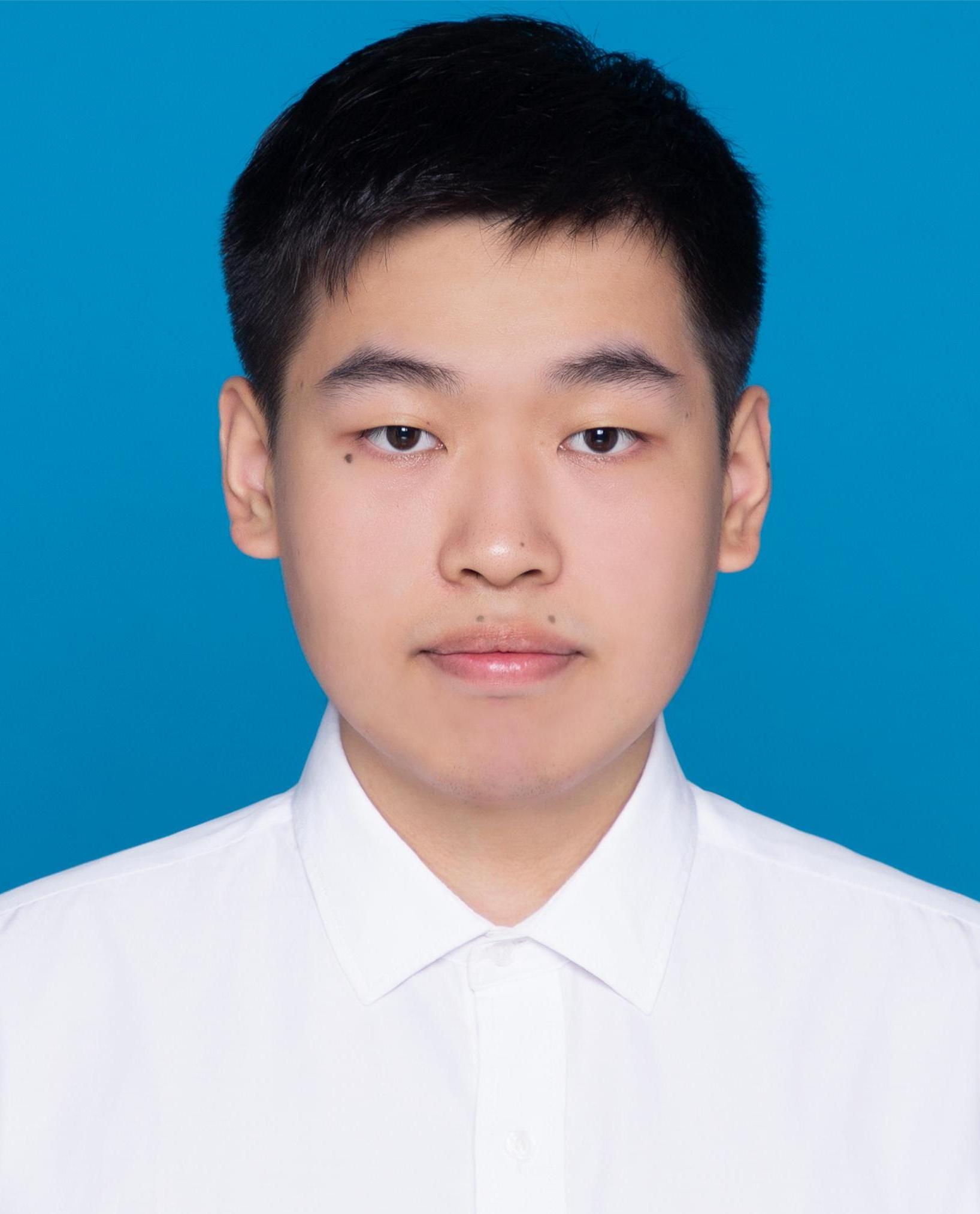}}]{Yingrui Wu}
is currently working toward the PhD degree at the Institute of Automation, Chinese Academy of Sciences (CAS). He received the Bachelor’s degree from Soochow University in 2023. His research interests include computer graphics and learning-based geometry
processing.
\end{IEEEbiography}

\vskip -1cm
\begin{IEEEbiography}
[{\includegraphics[width=1in,height=1.25in,clip,keepaspectratio]{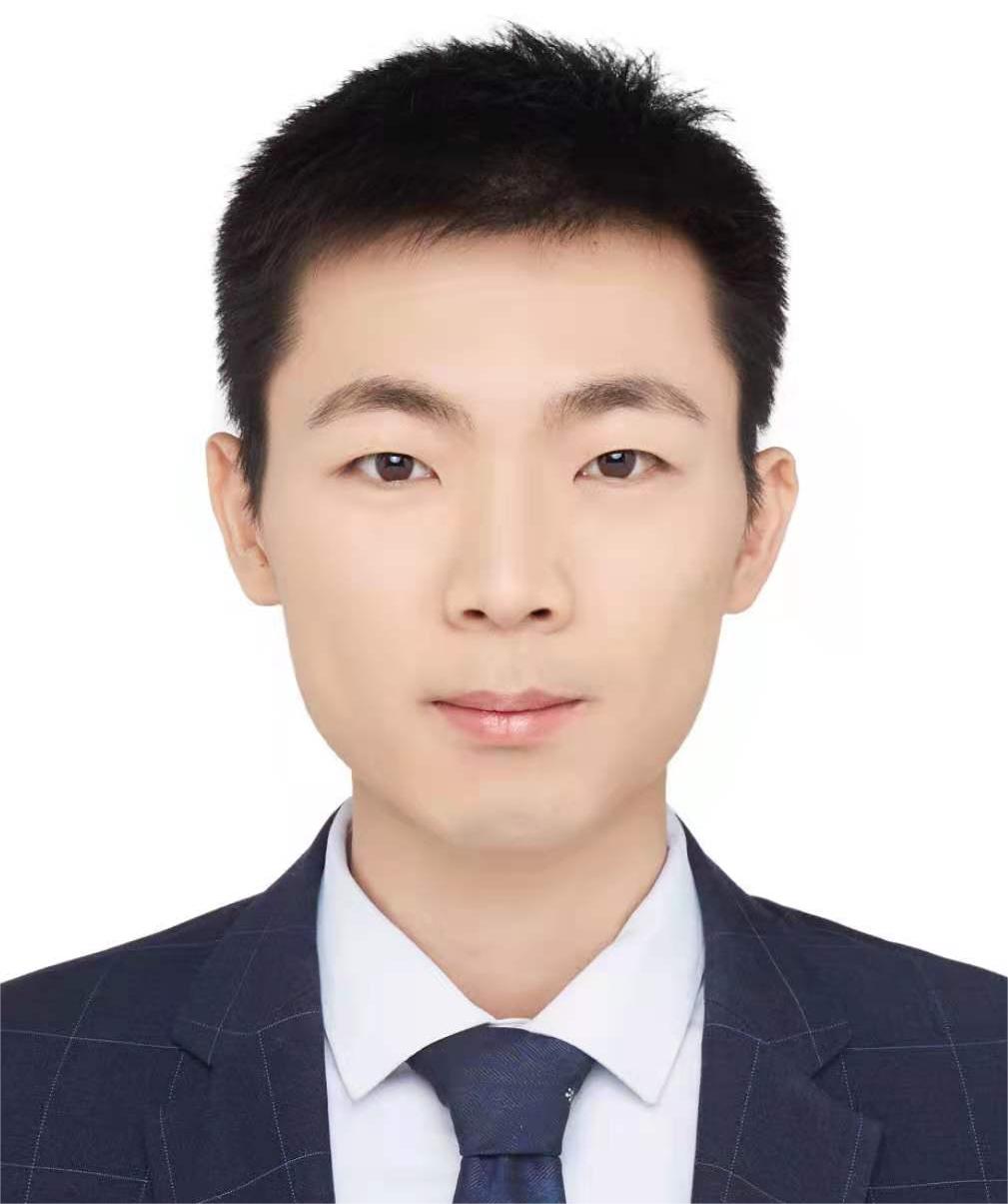}}]{Mingyang Zhao} is currently an assistant professor at the Hong Kong Institute of Science \& Innovation, CAS. He
 received the PhD degree from the Key Laboratory of Mathematics Mechanization, Academy of Mathematics and Systems Science, CAS, in
 2021. He now focuses his research and development on robust estimation, geometric problems, and shape analysis. He is awarded the grand Phd scholarship of Saudi Arabian and the excellent student prize of president fellowships, Chinese Academy of Sciences in 2020 and 2021, respectively. 
\end{IEEEbiography}

\begin{IEEEbiography}
[{\includegraphics[width=1in,height=1.25in,clip,keepaspectratio]{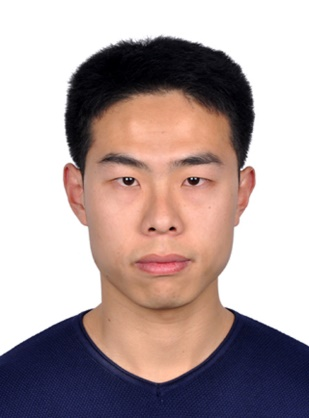}}] {Weize Quan} is currently an associate professor at the State Key Laboratory of Multimodal Artificial Intelligence Systems (MAIS), Institute of Automation, CAS. He received his Ph.D. degree from Institute of Automation, Chinese Academy of Sciences (China) and Universit\'e Grenoble Alpes (France) in 2020, and his Bachelor's degree from Wuhan University of Technology in 2014. His research interests include image processing, deep learning, and geometry processing.
\end{IEEEbiography}

\begin{IEEEbiography}
[{\includegraphics[width=1in,height=1.25in,clip,keepaspectratio]{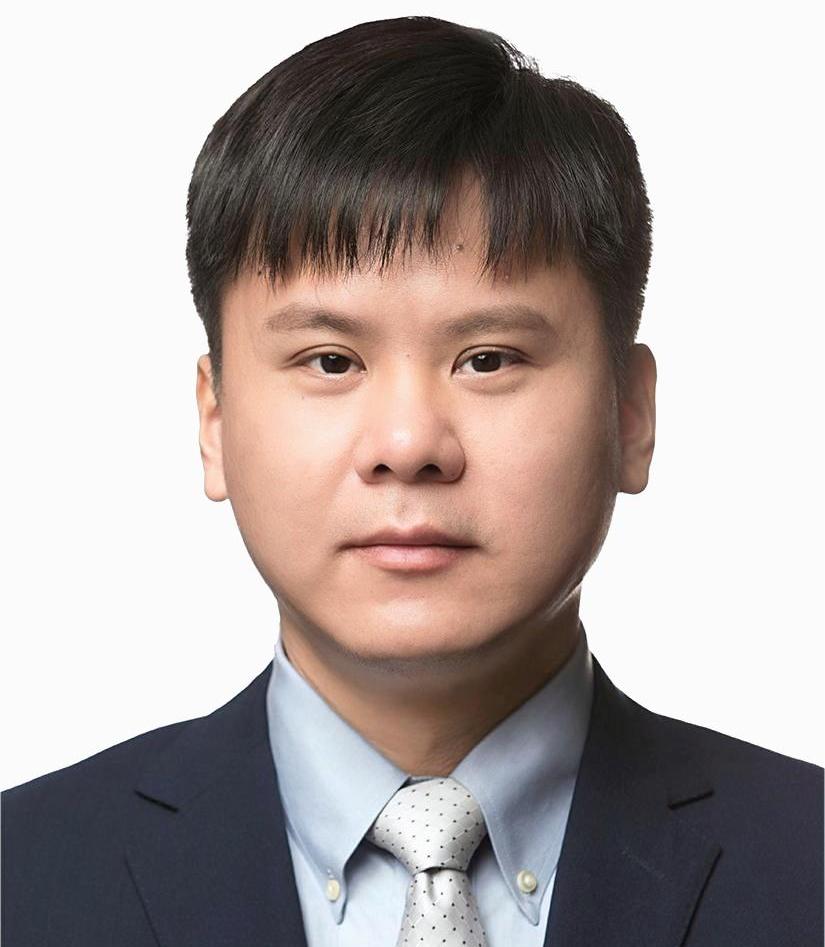}}] {Jian Shi} is an associate researcher at Institute of Automation, Chinese Academy of Sciences. His research interests include computer graphics and virtual reality.
\end{IEEEbiography}

\begin{IEEEbiography}[{\includegraphics[width=1in,height=1.25in,clip,keepaspectratio]{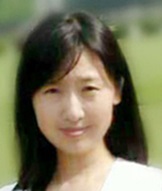}}]{Xiaohong Jia} is a professor at Key Laboratory of Mathematics Mechanization, Academy of Mathematics and Systems Science, CAS. She received her Ph.D.\ and Bachelor's degree from the University of Science and Technology of China in 2009 and 2004, respectively. Her research interests include computer graphics, computer aided geometric design and computational algebraic geometry.
\end{IEEEbiography}

\begin{IEEEbiography}[{\includegraphics[width=1in,height=1.25in,clip,keepaspectratio]{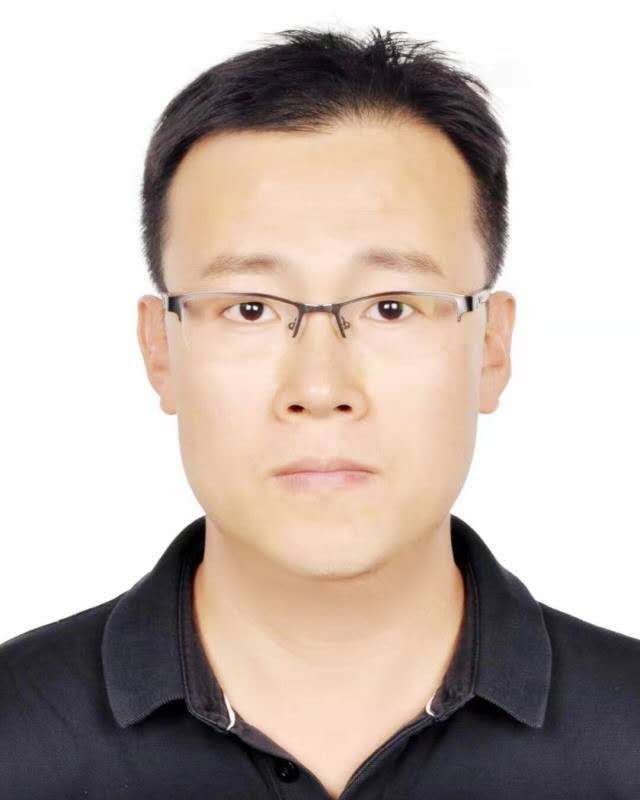}}]{Dong-Ming Yan} is a professor at the National Laboratory of Pattern Recognition of the Institute of Automation, CAS. He received his Ph.D.\ from Hong Kong University in 2010 and his Master's and Bachelor's degrees from Tsinghua University in 2005 and 2002, respectively. His research interests include computer graphics, computer vision, geometric processing and pattern recognition.
\end{IEEEbiography}

\end{document}